\documentclass[10pt,journal,compsoc]{IEEEtran}
\ifCLASSOPTIONcompsoc
  \usepackage[nocompress]{cite}
\else
  \usepackage{cite}
\fi
\usepackage{times}
\usepackage{epsfig}
\usepackage{graphicx}
\usepackage{amsmath}
\usepackage{amssymb}
\graphicspath{{figs/}}
\usepackage[utf8x]{inputenc} 
\usepackage[T1]{fontenc}    
\usepackage{url}            
\usepackage{booktabs}       
\usepackage{amsfonts}       
\usepackage{nicefrac}       
\usepackage{microtype}      
\usepackage{graphicx}
\usepackage{subcaption}
\usepackage{amsthm}
\usepackage{mathrsfs}
\usepackage{bm}
\usepackage{multirow}
\usepackage{multicol}
\usepackage{slashbox}

\theoremstyle{plain}

\ifCLASSINFOpdf
\else
\fi

\usepackage[pagebackref=true,breaklinks=true,letterpaper=true,colorlinks,bookmarks=false]{hyperref}

\usepackage{algorithm}
\usepackage[noend]{algpseudocode}
\begin{document}
%
\title{Deep Regionlets: Blended Representation and Deep Learning for Generic Object Detection}
%
%
%
%

\author{Hongyu~Xu,
		    Xutao~Lv,
        Xiaoyu~Wang,  
        Zhou~Ren,
        Navaneeth~Bodla, \\
        and~Rama~Chellappa,~\IEEEmembership{Fellow,~IEEE}
\IEEEcompsocitemizethanks{\IEEEcompsocthanksitem H. Xu is with Apple Inc., Cupertino, CA, 95014, USA. \protect\\
E-mail: {xuhongyu2006}@gmail.com
\IEEEcompsocthanksitem X. Lv and X. Wang are with Intellifusion. \protect\\
E-mail: \{lvxutao, fanghuaxue\}@gmail.com
\IEEEcompsocthanksitem Z. Ren is with Wormpex AI Research, Bellevue, WA, 98004, USA \protect\\
E-mail: {renzhou200622}@gmail.com
\IEEEcompsocthanksitem N. Bodla and R. Chellappa are with the Department
of Electrical and Computer Engineering, UMIACS, University of Maryland, College Park,
MD, 20742, USA. \protect\\
E-mail: \{nbodla, rama\}@umiacs.umd.edu}
}

\IEEEtitleabstractindextext{%

\begin{abstract}

In this paper, we propose a novel object detection algorithm named "Deep Regionlets" by integrating deep neural networks and a conventional detection schema for accurate generic object detection. Motivated by the effectiveness of regionlets for modeling object deformations and multiple aspect ratios, we incorporate regionlets into an end-to-end trainable deep learning framework. The deep regionlets framework consists of a region selection network and a deep regionlet learning module. Specifically, given a detection bounding box proposal, the region selection network provides guidance on where to select sub-regions from which features can be learned from. An object proposal typically contains 3-16 sub-regions. The regionlet learning module focuses on local feature selection and transformations to alleviate the effects of appearance variations. To this end, we first realize non-rectangular region selection within the detection framework to accommodate variations in object appearance. Moreover, we design a ``gating network" within the regionlet leaning module to enable instance dependent soft feature selection and pooling. The Deep Regionlets framework is trained end-to-end without additional efforts. We present ablation studies and extensive experiments on the PASCAL VOC dataset and the Microsoft COCO dataset. The proposed method yields competitive performance over state-of-the-art algorithms, such as RetinaNet and Mask R-CNN, even without additional segmentation labels. 

\end{abstract}

}

\maketitle

\IEEEdisplaynontitleabstractindextext

%
\IEEEpeerreviewmaketitle

\IEEEraisesectionheading{\section{Introduction}\label{sec:introduction}}



\IEEEPARstart{G}{eneric} object detection has been extensively studied in the computer vision community over several decades~\cite{Huang_HRSZKFFWSGM_CVPR2017,Bodla_BSCD_ICCV2017,Wang_WYZL_2015TPAMI,Girshick_GDDM_2014CVPR,Girshick_G_ICCV2015,Ren_RHGS_TPAMI2016,Dai_DLHS_NIPS2016,Lin_LGGHD_ICCV2017,Wang_WYZL_ICCV2013,Viola_VJ_CVPR2001,Dalal_DT_CVPR2005,Felzenszwalb_FGMR_TPAMI2010,Wang_WHY_ICCV2009,Zhang_ZWBLL_CVPR2018,Cai_CV_CVPR2018,Zhang_ZQXSWY_CVPR2018,Li_LSNWD_2019} due to its appeal to both academic research explorations as well as commercial applications. Object detection is also critical to many computer vision tasks such as scene understanding~\cite{Li_LSL_CVPR2009}, face recognition~\cite{Zhao_ZCPR_ACMCS2003,Turaga_TCSU_TCSVT2008,Ranjan_RBXSCCC_2018,Ranjan_RBZXGLNCCC_2018,Bodla_BZXCCC_WACV2017}, action recognition~\cite{Poppe_IVC2010,Jhuang_JGZSB_ICCV2013}, robotics and self-driving vehicles, etc. While designing object detection algorithms, two key challenges need to be carefully addressed: where the candidate locations are and how to discern whether they are the objects of interest. Although studied over several decades, accurate detection is still highly challenging due to cluttered backgrounds, occlusions, variations in object scale, pose, viewpoint and even part deformations. 



Prior works on object detection before the deep learning era addressed object deformations using various strategies based on hand-crafted features (\textit{i.e.} histogram of oriented gradients (HOG)~\cite{Dalal_DT_CVPR2005}, local binary pattern (LBP)~\cite{Ahonen_AHP_ECCV2004}, HOG-LBP~\cite{Wang_WHY_ICCV2009}, scale-invariant feature transform (SIFT)~\cite{Lowe_L_ICCV1999}), etc. One of the earliest works, the classical Adaboost~\cite{Viola_VJ_IJCV2004} detector deployed an ensemble classifier of fast features to model local variations, especially for the detection of faces or pedestrians. The deformable part model (DPM)~\cite{Felzenszwalb_FGMR_TPAMI2010} first proposed to model object deformations explicitly using latent variables, improved localization precision. However, these approaches usually involve exhaustive search for possible locations, scales and aspect ratios of the object, by using the sliding window approach. Furthermore, spatial pyramid matching of bag-of-words (BoW) models~\cite{Dollar_DAK_ECCV2012} in image classification, has been adopted for object detection, providing robustness to large deformations. The computational cost has been alleviated by using thousands of object-independent bounding box proposals.  

Owing to its ability to efficiently learn a descriptive and flexible object representation, Wang \textit{et al}.'s regionlet-based detection framework~\cite{Wang_WYZL_ICCV2013} has gained a lot of attention. It provides the flexibility to deal with different scales and aspect ratios without performing exhaustive search. It first proposed the concept of \textbf{\emph{regionlet}} by defining a three-level structural relationship: candidate bounding boxes (sliding windows), regions inside the bounding box and groups of regionlets (sub-regions inside each region). It operates by directly extracting features from regionlets in several selected regions within an arbitrary detection bounding box and performs (max) pooling among the regionlets. Such a feature extraction hierarchy is capable of dealing with variable aspect ratios and  flexible feature sets, which leads to improved learning of robust feature representation of the object for region-based object detection. 

\begin{figure*}[]
\centering{
\includegraphics[width=1.9\columnwidth]{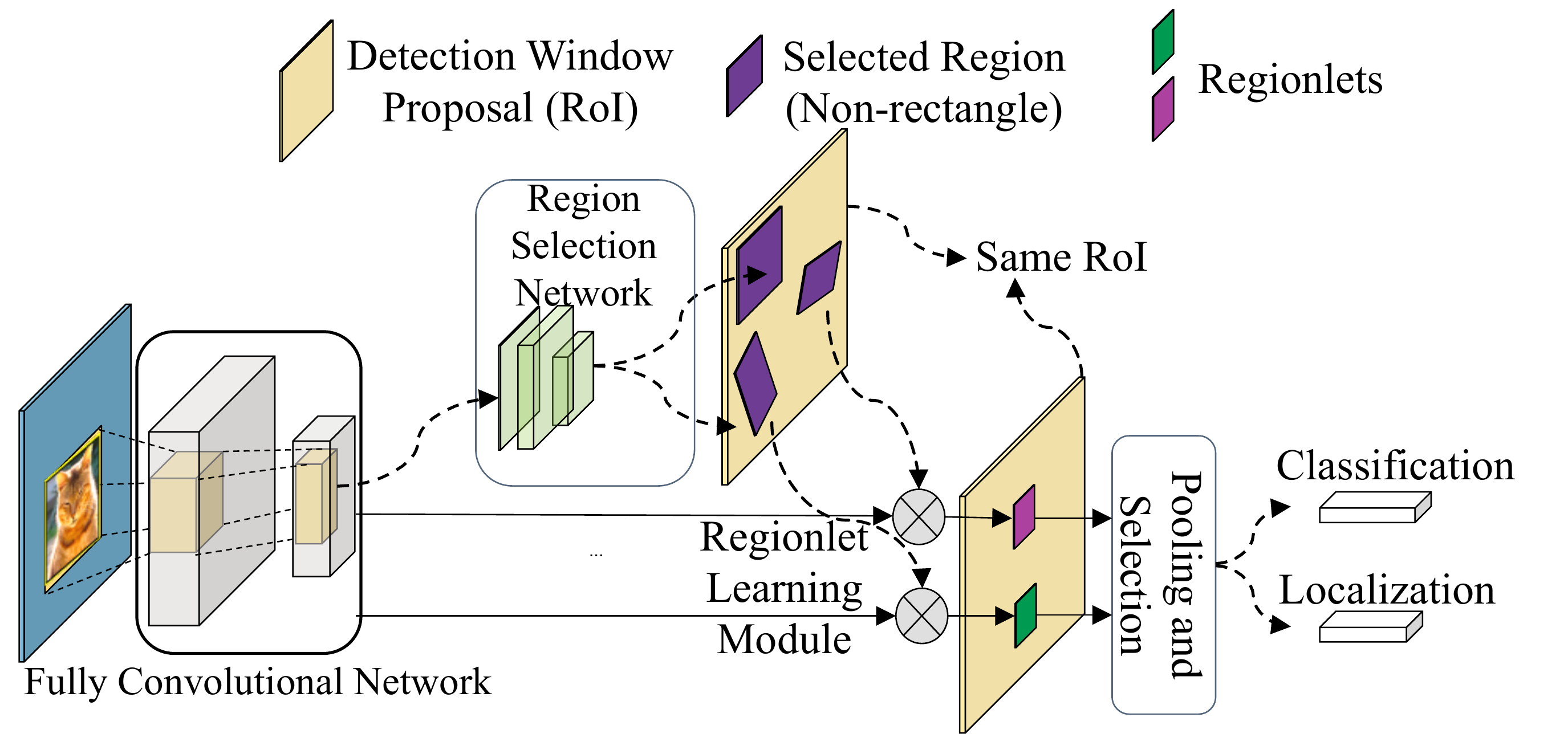}}
\caption{Architecture of the Deep Regionlets detection approach. It consists of a Region Selection Network (RSN) and a deep regionlet learning module. A region is one part of the detection window. The region selection network performs \emph{non-rectangular} region selection from the detection window proposal generated by the object proposal network. The deep regionlet learning module learns the feature selection schema (corresponding to functionality of the original regionlet) through a spatial transformation network and a gating network. The entire pipeline is end-to-end trainable. For clarity of visualization, object proposal network is not displayed here. }
\label{fig::DeepRegionlet_overall}
\end{figure*}


Recently, deep learning has achieved significant success on many computer vision tasks such as image classification~\cite{Krizhevsky_KSH_2012NIPS,He_HZRS_CVPR16,Jia_JEJSJRST_2014,Wei_WXLHNDZY_PAMI2016}, semantic segmentation~\cite{Long_LSD_CVPR2015} and object detection~\cite{Girshick_GDDM_2014CVPR,Girshick_G_ICCV2015} 
Despite the superior performance of deep learning-based detection approaches, most network architectures~\cite{Ren_RHGS_TPAMI2016,Dai_DLHS_NIPS2016,Liu_LAESRFB_CVPR2016} do not take advantage of successful conventional ideas such as DPM or regionlets. Those conventional methods have been shown to be effective for modeling object deformation, sub-categories and multiple aspect ratios. As
deep convolutional neural networks~\cite{Yan_YLYP_PIEEE1998} exhibit superior capability in learning hierarchical and discriminative features (deep features), it motivates us to think of how to "build" the bridge between deep neural networks and conventional object detection schemes.

Recent advances~\cite{Ouyang_OZWQLTLYWL_TPAMI2017,Dai_DQXLZHW_ICCV2017,Mordan_MTCH_2017} have answered this question by combining the conventional DPM-based detectors with deep neural network architectures and achieved promising results. Yet few methods~\cite{Zou_ZWSL_BMVC2014,Wang_WYZL_2015TPAMI} have been explored for conventional regionlet detection schema. Zou \textit{et al}.~\cite{Zou_ZWSL_BMVC2014} made the first attempt to utilize deep features instead of hand-crafted features. They introduced the dense neural pattern (DNP) to extract features from an image with arbitrary resolution using a well-trained deep convolutional neural network (\textit{i.e} AlexNet~\cite{Krizhevsky_KSH_2012NIPS}). Activations from the same receptive fields but different feature maps can be extracted by back-tracking to exact coordinates in the original image. Though~\cite{Zou_ZWSL_BMVC2014} presented effective performance boost with deep features, several limitations make DNP~\cite{Zou_ZWSL_BMVC2014} not applicable to deep neural networks developed recently (\textit{i.e.} VGG~\cite{Simonyan_SZ_2014}, ResNet~\cite{He_HZRS_CVPR16}). First, as DNP uses selective search to generate the region proposal, it would be extremely inefficient to extract the feature activations with more sophisticated deep neural networks. 
Second, end-to-end training is not feasible as DNP~\cite{Zou_ZWSL_BMVC2014} directly extracted features on well-trained deep neural network for classification tasks.


These observations motivate us to establish an improved bridge between deep convolutional neural networks and conventional \emph{regionlet} object detection schema. In this paper, we incorporate the conventional regionlet method into an \emph{end-to-end} trainable deep learning framework. Despite being able to handle arbitrary bounding boxes, several drawbacks arise when directly integrating the regionlet methodology into the deep learning framework. 
First,  both regionlet~\cite{Wang_WYZL_ICCV2013} and DNP~\cite{Zou_ZWSL_BMVC2014} proposed to learn cascade object classifiers after hand-crafted/deep feature extraction in each regionlet, thus end-to-end learning is not feasible in both frameworks. Second, object sub-regions in regionlet-based detection have to be rectangular, which does not effectively model object deformations. Moreover, both regions and regionlets are fixed after training is completed. 

In this paper, we propose a novel object detector named "\textbf{Deep Regionlets}" by blending deep learning and the traditional regionlet method~\cite{Wang_WYZL_ICCV2013,Zou_ZWSL_BMVC2014}. The proposed framework "Deep Regionlets" is able to address the limitations of both traditional regionlet and its DNP extension, leading to significant improvements in precision by exploiting the power of deep convolutional neural networks.  

The overall design of the proposed detection system is illustrated in Figure~\ref{fig::DeepRegionlet_overall}. It consists of a region selection network and a deep regionlet learning module. The region selection network (RSN) performs \emph{non-rectangular} region selection from the detection window proposal\footnote{The detection window proposal is generated by a region proposal network~\cite{Ren_RHGS_TPAMI2016,Dai_DLHS_NIPS2016,Girshick_G_ICCV2015}. It is also called region of interest (ROI)} (RoI) to address the limitations of the traditional regionlet approach.
We further design a deep regionlet learning module to learn the regionlets through a spatial transformation and a gating network. By using the proposed gating network, which is a soft regionlet selector, the final feature representation is more suitable for detection. The entire pipeline is end-to-end trainable using only the input images and ground truth bounding boxes.

We conduct a detailed analysis of our approach to understand its merits and properties. Extensive experiments on two detection benchmark datasets, PASCAL VOC~\cite{Everingham_EGWWZ_IJCV2010} and Microsoft COCO~\cite{Lin_LMBHPRDZ_ECCV2014} show that the proposed deep regionlet approach outperforms several competitors~\cite{Ren_RHGS_TPAMI2016,Dai_DLHS_NIPS2016,Dai_DQXLZHW_ICCV2017,Mordan_MTCH_2017,Zhang_ZWBLL_CVPR2018}. Even without segmentation labels, we outperform state-of-the-art algorithms Mask R-CNN~\cite{He_HGDG_ICCV2017} and RetinaNet~\cite{Lin_LGGHD_ICCV2017}. 

To summarize, the major contributions of this paper are: 
\begin{itemize}
\item We propose a novel approach for object detection, "Deep Regionlets". Our work blends the traditional regionlet method and the deep learning framework. The system could be trained in an end-to-end manner.
\item We design a region selection network, which \textbf{first} performs \textbf{non-rectangular} regions selection within the detection bounding box generated from a detection window proposal. It provides more flexibility for modeling objects with variable shapes and deformable parts.
\item We propose a deep regionlet learning module, including feature transformation and a gating network. The gating network serves as a soft regionlet selector and lets the network focus on features that benefit detection performance.
\item We present empirical results on object detection benchmark datasets, which demonstrates competitive performance over state-of-the-art.
\end{itemize}

A preliminary version of this work was published in~\cite{Xu_XLWRBC_ECCV2018}. In this paper, we extend~\cite{Xu_XLWRBC_ECCV2018} in the following aspects: I) we propose a new design for RSN and the deep regionlet learning module. In particular, we design the RSN to predict a set of \textit{projective} transformation parameters. By extending RSN from predicting the affine parameters~\cite{Xu_XLWRBC_ECCV2018} to projective parameters, we first realize non-rectangular (quadrilateral) region selection within the detection bounding box. 
II) we extensively study the behavior of the proposed approach, especially it's performance over traditional the regionlet detection scheme~\cite{Wang_WYZL_2015TPAMI,Zou_ZWSL_BMVC2014};
III) we further provide a theoretical analysis as well as more empirical support, which further demonstrates the effectiveness of the proposed deep regionlets approach;  IV) more experimental results are presented for performance evaluation.

The rest of the paper is organized as follows: Section~\ref{sec:relatedWork} briefly discusses both traditional object detection approaches and deep learning-based approaches. Section~\ref{sec:regionletsReview} reviews the traditional regionlet-based approach and its dense neural pattern.
Section~\ref{sec:method} presents the proposed deep regionlet approach. Section~\ref{sec:related_discussion} discusses both similarities with and differences from recent works. 
Section~\ref{sec:experiments} provides detailed experimental results and analysis on object detection benchmark datasets PASCAL VOC and MS COCO. Section~\ref{sec:conclusion} concludes the paper.

\section{Related Work}\label{sec:relatedWork}

Generic object detection accuracy has improved over years largely due to more effecive handling of multi-viewpoints, modeling deformations~\cite{Felzenszwalb_FGMR_TPAMI2010}, and the success of deep learning techniques~\cite{Girshick_GDDM_2014CVPR,Girshick_G_ICCV2015,Ren_RHGS_TPAMI2016,He_HZRS_CVPR16,Simonyan_SZ_2014}. A comprehensive survey of object detection literature is beyond the scope of this paper. 

Briefly speaking, many approaches based on traditional representations~\cite{Felzenszwalb_FGMR_TPAMI2010,Wang_WYZL_ICCV2013,Viola_VJ_CVPR2001} and deep learning~\cite{Girshick_G_ICCV2015,Ren_RHGS_TPAMI2016,Liu_LAESRFB_CVPR2016,Redmon_RDGF_CVPR2016,Dai_DLHS_NIPS2016,Girshick_GDDM_2014CVPR,He_HZRS_ECCV20014,Dai_DQXLZHW_ICCV2017,Mordan_MTCH_2017,Fu_FLRTB_2017,Zhang_ZQXSWY_CVPR2018,Zhang_ZWBLL_CVPR2018,Cai_CV_CVPR2018,Hu_HGZDW_CVPR2018,Cheng_CWSFXH_ECCV2018,Li_LPYZDS_ECCV2018,Jiang_JLMXJ_ECCV2018,Jiang_JLMXJ_ECCV2018,Wei_WPQOY_ECCV2018} have been proposed. Traditional approaches mainly used hand-crafted features (\textit{i.e.} LBP~\cite{Ahonen_AHP_ECCV2004}, HOG)~\cite{Dalal_DT_CVPR2005}) to design object detectors using the sliding window paradigm. One of the earliest works~\cite{Viola_VJ_CVPR2001} used boosted cascaded detectors for face detection, which led to its wide adoption. Deformable part models~\cite{Felzenszwalb_FGM_CVPR2010} further extended the cascaded detectors to more general object categories. Due to rapid development of deep learning techniques~\cite{Krizhevsky_KSH_2012NIPS,He_HZRS_CVPR16,Simonyan_SZ_2014}, the deep learning-based detectors have become dominant object detectors. 

Deep learning-based detectors could be further categorized into two classes, single-stage detectors and two-stage detectors, based on whether the detectors have proposal-driven mechanism or not. The single-stage detectors~\cite{Sermanet_SEZMFL_ICLR2014,Redmon_RDGF_CVPR2016,Liu_LAESRFB_CVPR2016,Fu_FLRTB_2017,Lin_LDGHHB_CVPR2017,Lin_LGGHD_ICCV2017,Zhang_ZQXSWY_CVPR2018,Zhang_ZWBLL_CVPR2018,Kong_KSHL_ECCV2018,Kim_KKSKK_ECCV2018,Law_LD_ECCV2018} apply regular, dense sampling windows over object locations, scales and aspect ratios. By exploiting multiple layers within a deep CNN network directly, the single-stage detectors are faster but their detection accuracy is often low compared to two-stage detectors. 

Two-stage detectors~\cite{Girshick_G_ICCV2015,Ren_RHGS_TPAMI2016,Dai_DLHS_NIPS2016,Dai_DQXLZHW_ICCV2017,Mordan_MTCH_2017,He_HGDG_ICCV2017,Cheng_CWSFXH_ECCV2018,Cheng_CWSFXH_2018,Chen_CHT_ECCV2018,Liu_LHW_ECCV2018,Gu_GHWWD_ECCV2018,Zhu_ZZWZWL_ICCV2017,Singh_SD_CVPR2018,Singh_SND_2018,Wu_WSNXKW_ICCV2019} first generate a sparse set of candidate proposals of detection bounding boxes by the region proposal network. After filtering out the majority of negative background boxes by RPN, the second stage classifies the detection bounding box proposals and performs regression to predict the object categories and their corresponding locations. Cheng \textit{et al}.~\cite{Cheng_CWSFXH_ECCV2018} propose a decoupled classification refinement approach which can effectively ease the false positive samples. The two-stage detectors consistently achieve higher accuracy than single-stage detectors and numerous extensions have been proposed~\cite{Girshick_G_ICCV2015,Ren_RHGS_TPAMI2016,Dai_DLHS_NIPS2016,Dai_DQXLZHW_ICCV2017,Mordan_MTCH_2017,He_HGDG_ICCV2017}. Our method follows the two-stage detector architecture by taking advantage of RPN without the need for dense sampling of object locations, scales and aspect ratios. 

\section{Traditional Regionlets for Detection}
\label{sec:regionletsReview}

In this section, we review traditional regionlet-based approach and its dense neural pattern extension as our work is directly motivated by the regionlet detection scheme. We incorporate \emph{regionlet} into an end-to-end trainable deep learning framework. The proposed deep regionlets framework overcomes the limitations of both traditional regionlet method~\cite{Wang_WYZL_ICCV2013} and DNP~\cite{Zou_ZWSL_BMVC2014}, leading to significant improvement in detection performance. 

\subsection{Regionlets Definition}

\begin{figure}[!tbh]
\centering{
\includegraphics[width=0.7\columnwidth]{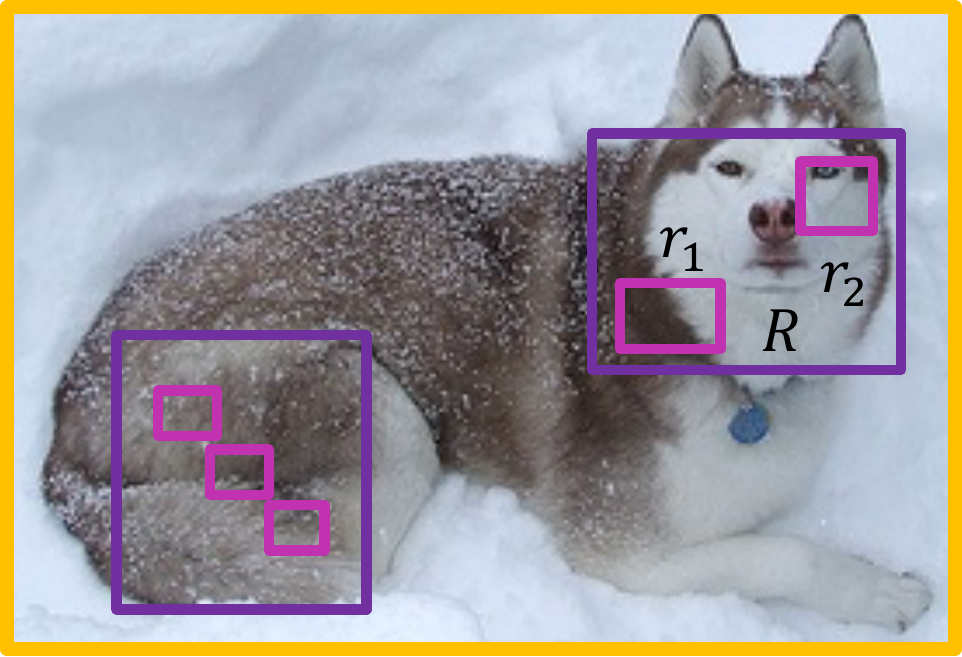}}
\caption{Illustration of structural relationships among the detection bounding box, feature extraction regions and regionlets. The yellow box is a detection bounding box and $R$ is a feature extraction region shown as a purple rectangle with filled dots inside the bounding box. Inside $R$, two small sub-regions denoted as $r_1$ and $r_2$ are form \emph{regionlets}.} 
\label{fig::regionlet_illustration}
\end{figure}

A \emph{regionlet} is a base feature extraction region defined proportionally with respect to a sliding window or a detection bounding box) at arbitrary resolution (\textit{i.e.} size and aspect ratio). Wang \textit{et al.}~\cite{Wang_WYZL_ICCV2013,Wang_WYZL_2015TPAMI} first introduced the concept of regionlets, illustrated in Figure~\ref{fig::regionlet_illustration}. 

Figure~\ref{fig::regionlet_illustration} shows a three-level structure consisting of a detection bounding box, number of regions inside the bounding box and a group of regionlets (sub-regions inside each region). In Figure~\ref{fig::regionlet_illustration}, the yellow box is a detection bounding box.  
A rectangular feature extraction region inside the detection bounding box is denoted as $R$ (purple rectangle). Furthermore, within this region $R$, we spot some small sub-regions (small magenta rectangles) $r_{i \{i = 1\dots N\}}$(\textit{e.g.} $r_1$, $r_2$) and define them as a set of \emph{regionlets}. One of the motivations behind the term \emph{regionlet} is that the hand-crafted features extracted from these sub-regions will be aggregated into a single feature representation of $R$. 

To summarize, a detection bounding box is represented by several regions, each of which consists of a small set of regionlets. 
By using the \emph{relative} positions and sizes of regionlets and regions, the difficulty of the arbitrary detection bounding box has been well addressed. Therefore, the regionlet-based representation is able to model relative spatial layouts inside an object and can be efficiently applied to arbitrary bounding boxes at different scales and aspect ratios. However, the initialization of regionlets possess randomness and both regions ($R$) and regionlets (\textit{i.e.} $r_1$, $r_2$) are fixed after training is completed. Moreover, it is based on hand-crafted features (\textit{i.e.} HOG~\cite{Dalal_DT_CVPR2005} or LBP descriptors~\cite{Ahonen_AHP_ECCV2004}) in each regionlet respectively. The proposed deep regionlet-based approach in Section~\ref{sec:method} mitigates such limitations.

\subsection{Dense Neural Pattern Extension}
Despite the success of the sophisticated regionlet detection method~\cite{Wang_WYZL_ICCV2013}, the features employed are still hand-crafted representations such as LBP~\cite{Ahonen_AHP_ECCV2004}, HOG~\cite{Dalal_DT_CVPR2005} or covariance-based on the gradients of the image. 

As with the significant success of deep learning, the deep convolutional neural network has demonstrated promising performance on object detection~\cite{Girshick_GDDM_2014CVPR,Girshick_G_ICCV2015}. The dramatic improvements are due to hierarchically learned complex features (deep features) from deep neural network. One intuitive way to improve the traditional regionlet-based approach is to utilize deep features instead of hand-crafted features. Zou~\textit{et al.}~\cite{Zou_ZWSL_BMVC2014} made the first attempt to incorporate the regionlet detection scheme in the deep neural network by introducing DNP that extracts features from an image with arbitrary resolution using a well trained classification-based deep convoluttional neural network (\textit{i.e.} AlexNet~\cite{Krizhevsky_KSH_2012NIPS}). 

However, there are several limitations which make it challenging to integrate DNP~\cite{Zou_ZWSL_BMVC2014} with recently developed deep neural networks (\textit{i.e.} VGG~\cite{Simonyan_SZ_2014}, ResNet~\cite{He_HZRS_CVPR16}). First, DNP~\cite{Zou_ZWSL_BMVC2014} directly extracted features from well-trained deep neural network for the classification task, hence end-to-end training is not feasible. Second, it used the selective search to generate region proposals. The object proposal approach is a generative method which uses bottom up image segmentation results to create bounding box proposals. As this step heavily relies on image edges, thousands or even tens of thousand object proposals are needed when the image quality is bad, which makes it inferior to supervised methods such as region proposal networks.  The proposed deep regionlet-based approach overcomes drawbacks of both the traditional regionlet method and DNP, leading to significant improvements in detection accuracy.

\section{Deep Regionlets}
\label{sec:method}

In this section, We first present the overall design of the proposed deep regionlet approach with end-to-end training and then describe each module in detail.


\subsection{System Architecture}

Modern deep learning-based object detection methods typically consists of several critical parts, object proposal generation networks (such as the region proposal network approach, noted as RPN), object classification networks and object coordinate regression networks.  
 RPN generates a set of candidate bounding boxes\footnote{~\cite{Ren_RHGS_TPAMI2016,Dai_DLHS_NIPS2016,Girshick_G_ICCV2015} also called the detection bounding box as detection window proposal.}.  Subsequently, the classification networks determine whether the bounding box contains the object of interest by pooling the convolutional features inside the proposed bounding box generated by the RPN~\cite{Ren_RHGS_TPAMI2016,Dai_DLHS_NIPS2016} and performing a linear classification. Taking advantage of this detection network, we introduce the overall design of the proposed object detection framework, named "Deep Regionlets", as illustrated in Figure~\ref{fig::DeepRegionlet_overall}. 

The general architecture consists of a Region Selection Network (or sub-Region Selection Network, short for RSN) and a deep regionlet learning module. RSN is performed following the RPN. The RPN generates object proposals, while the RSN selects sub-regions inside the object proposal for flexible feature extraction. In particular, RSN is used to predict transformation parameters to select regions given a candidate bounding box, which is generated by the RPN. The regionlets are further learned within each selected region defined by the RSN. The system is designed to be trained in a fully end-to-end manner using only input images and the ground truth bounding box. The RSN as well as the regionlet learning module can be simultaneously learned over each selected region given the detection bounding box proposal. We will interchangeably refer to sub-Region Selection Network and RSN based on the context without confusion.

\subsection{Region Selection Network \label{sec::region_selection}}

We design the RSN to have the following properties: 
\begin{itemize}
	\item End-to-end trainable.
	\item Possess simple structure without introducing too many parameters.
	\item Generate regions with arbitrary shapes.
\end{itemize}

Keeping these in mind, we design the RSN to predict a set of \emph{projective} transformation parameters. 
By using these projective transformation parameters, as well as not requiring the regions to be rectangular, we have more flexibility in modeling an object with arbitrary shape and deformable parts. 

Specifically, we design the RSN using a small neural network with three fully connected layers. The first two fully connected layers have output size of $256$, with ReLU activations. The last fully connected layer has the output size of nine, which is used to predict the set of projective transformation parameters $\Theta = [ \theta_1, \theta_2, \theta_3; \theta_4, \theta_5, \theta_6; \theta_7, \theta_8, \theta_9 ]$. It is noted that in the preliminary version~\cite{Xu_XLWRBC_ECCV2018}, RSN is only designed to predict the set of affine parameters $\Theta^\ast = [ \theta_1, \theta_2, \theta_3; \theta_4, \theta_5, \theta_6]$

Note that the candidate detection bounding boxes proposed by RPN have arbitrary sizes and aspect ratios. In order to address this difficulty, we use \emph{relative} positions and sizes of the selected region within a detection bounding box. The candidate bounding box generated by RPN is defined by the top-left point ($w_0, h_0$), width $w$ and height $h$ of the box. We normalize the coordinates by the width $w$ and height $h$. As a result, we could use the normalized projective transformation parameters $\Theta = [ \theta_1, \theta_2, \theta_3; \theta_4, \theta_5, \theta_6; \theta_7, \theta_8, \theta_9]$ ($\theta_i \in [-1, 1]$) to evaluate one selected region within one candidate detection window at different sizes and aspect ratios without scaling images into multiple resolutions or using multiple-components to enumerate possible aspect ratios, like anchors~\cite{Ren_RHGS_TPAMI2016,Liu_LAESRFB_CVPR2016,Fu_FLRTB_2017}. \newline 

\subsubsection{Initialization of sub-Region Selection Network}
Taking advantage of the \emph{relative} and \emph{normalized} coordinates, we initialize the RSN by equally dividing the whole detecting bounding box to several sub-regions, named as \emph{cell}s, without any overlap among them. 

\begin{figure}[!bth]
\centering{
\includegraphics[width=0.95\columnwidth]{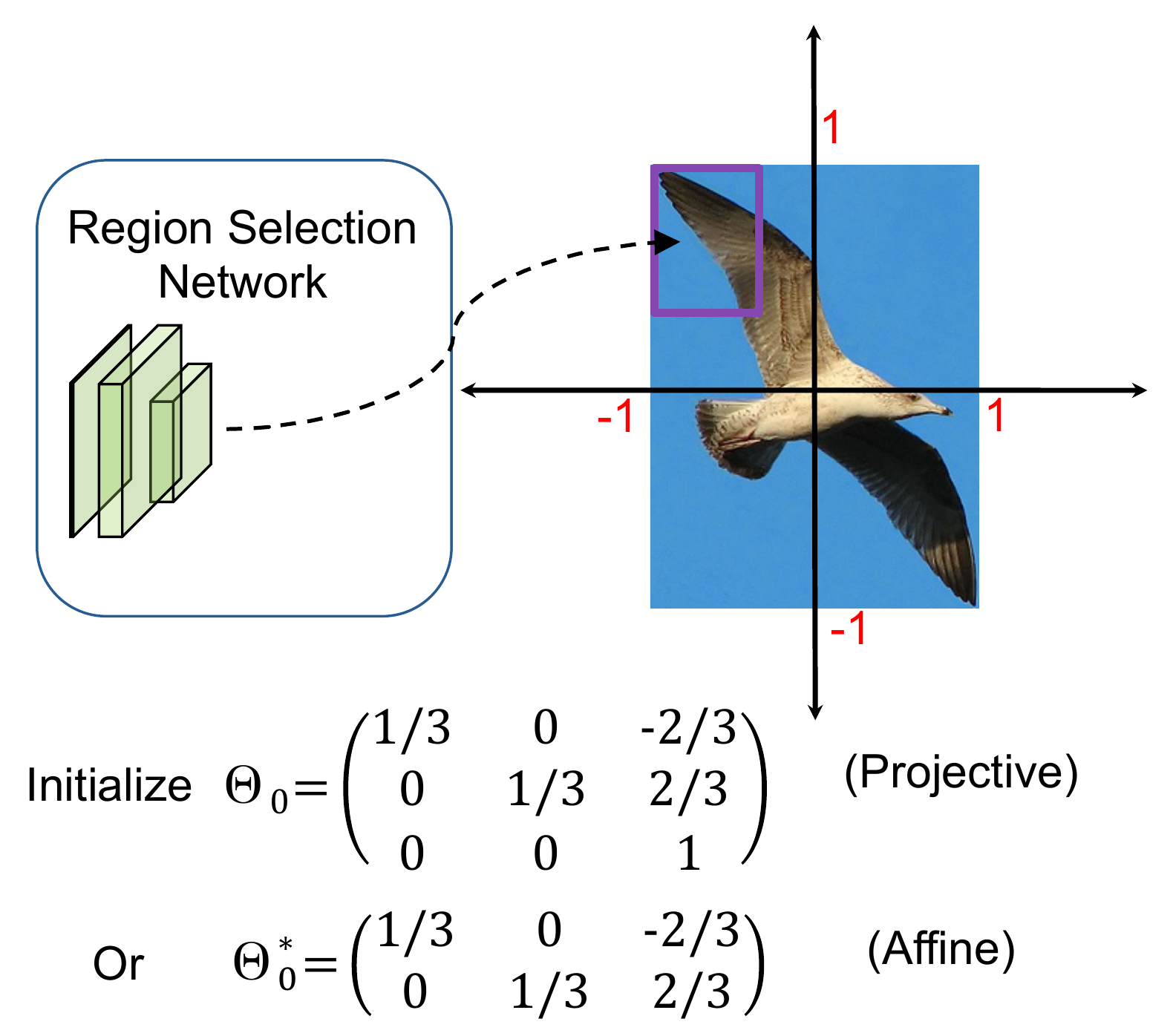}}
\caption{Initialization of one set of projective transformation parameters and affine transformation parameters.  Normalized projective transformation parameters $\Theta_0 = [\frac{1}{3}, 0, -\frac{2}{3}; 0, \frac{1}{3}, \frac{2}{3}; 0, 0, 1]$ ($\theta_i \in [-1, 1]$) and affine transformation parameters $\Theta_0^\ast= [\frac{1}{3}, 0, -\frac{2}{3}; 0, \frac{1}{3}, \frac{2}{3}]$ ($\theta_i^\ast \in [-1, 1]$) selects the top-left region in the $3\times 3$ evenly divided detection bounding box, shown as the purple rectangle.}
\label{fig::rsn_parameter_initialization}
\end{figure}

Figure~\ref{fig::rsn_parameter_initialization} shows an example of initialization from one projective transformation and one affine transformation in~\cite{Xu_XLWRBC_ECCV2018}. 
(\textit{i.e.} $3\times3$). The first cell, which is the top-left bin in the whole region (detection bounding box) could be defined by initializing the corresponding projective transformation parameters $\Theta_0 = [\frac{1}{3}, 0, -\frac{2}{3}; 0, \frac{1}{3}, \frac{2}{3}; 0, 0, 1]$ or affine transformation parameters $\Theta_0^\ast= [\frac{1}{3}, 0, -\frac{2}{3}; 0, \frac{1}{3}, \frac{2}{3}]$. The other eight of $3 \times 3$ cells are initialized in a similar way.

\subsection{Deep Regionlet Learning \label{sec::stn_regionlet}}

The original Regionlet work further divides object sub-regions into smaller rectangular areas (named as regionlets) and performs hard feature selection and pooling among these areas. Our approach extends this idea by performing the fine feature selection and pooling in a soft way using a gating network. After object sub-regions are selected by RSN, regionlets are further learned from the selected region defined by the normalized projective (affine) transformation parameters. Note that our motivation is to design the network to be trained in a fully end-to-end manner using only input images and ground truth bounding boxes. Therefore, both the selected regions and regionlet learning should be able to be trained by deep neural networks. Moreover, we would like the regionlets extracted from the selected regions to better represent objects with variable shapes and deformable parts.

Inspired by the spatial transform network~\cite{Jaderberg_JSZK_NIPS2015,Zhong_ZCH_SPL2017}, any parameterizable transformation including translation, scaling, rotation, affine or even projective transformation can be learned by a spatial transformer. In this section, we introduce our deep regionlet learning module to learn the regionlets in the selected region, which is defined by the projective transformation parameters. 

More specifically, we aim to learn regionlets from one selected region defined by one set of projective transformation $\Theta$ to better match the shapes of objects. This is done with a selected region $R$ from RPN, transformation parameters $\Theta = [ \theta_1, \theta_2, \theta_3; \theta_4, \theta_5, \theta_6; \theta_7, \theta_8, \theta_9]$ predicted by RSN and a set of feature maps $U = \{U_{i}, i = 1,\dots,n \}$. Without loss of generality, let $U_{i}$ be one of the feature map out of the $n$ feature maps. A selected region $R$ is of size $w \times h$ with the top-left corner $(w_0, h_0)$. Inside the $U_{i}$ feature map, we present the regionlet learning module as follows:

Let $s$ denote the source and $t$ denote target, we define $(x_p^s, y_p^s)$ as the $p$-th spatial location in the original feature map $U_{i}$ and $(x_p^s, y_p^s)$ as the corresponding spatial location in the output feature maps after projective transformation. First, a grid generator~\cite{Jaderberg_JSZK_NIPS2015} generates the source map coordinates ($x_p^s, y_p^s, 1$) based on the transformation parameters, given the $p$-th spatial location ($x_p^t, y_p^t, 1$) in target feature maps. The process is generated using (\ref{eqn::grid_generator}) given below. 

\begin{equation}
\begin{split}
\begin{pmatrix} x_{p}^s \\ y_{p}^s \\ 1 \end{pmatrix} = \frac{1}{z_p^{s}} 
\begin{pmatrix}
\theta_1 \quad \theta_2 \quad \theta_3 \\
\theta_4 \quad \theta_5 \quad \theta_6 \\
\theta_7 \quad \theta_8 \quad 1
\end{pmatrix} 
\begin{pmatrix} x_{p}^t \\ y_{p}^t \\ 1
\end{pmatrix} 
\label{eqn::grid_generator}
\end{split}
\end{equation}
where $\theta_9 = 1 $ and $z_p^{s} = \theta_7 x_p^t + \theta_8 y_p^t + 1$ to ensure the third dimension of ($x_p^s, y_p^s, 1$) is $1$.

Next, the sampler samples the input feature $U$ at the generated source coordinates. Let $U_{nm}^c$ be the value at location $(n, m)$ in channel $c$ of the input feature. The total output feature map $V$ is of size $H \times W$. $V(x_p^t, y_p^t, c| \Theta,R)$ be the output feature value at location ($x_p^t, y_p^t$) ($x_p^t\in [0, H]$, $y_p^t\in [0, W]$) in channel $c$, which is computed as

\begin{equation}
\begin{split}
V(x_p^t, y_p^t, c| \Theta,R) = \sum_{n}^{H} & \sum_{m}^{M}  U_{nm}^c \max(0, 1 - | x_p^s - m|) \\ 
& \max(0, 1 - | y_p^s - n|) 
\label{eqn::stroi_bilinear}
\end{split}
\end{equation}

\subsubsection{Back Propagation through Spatial Transform}
To allow back propagation of the loss through the regionlet learning module, we can define the gradients with respect to the feature maps and the region selection network. In this layer's \texttt{backward} function, we have partial derivative of the loss function with respect to feature map variable $U_{mn}^c$ and projective transform parameter $\Theta = [ \theta_1, \theta_2, \theta_3; \theta_4, \theta_5, \theta_6; \theta_7, \theta_8, \theta_9]$. Motivated by~\cite{Jaderberg_JSZK_NIPS2015}, the partial derivative of the loss function with respect to the feature map is:

\begin{equation}
\begin{split}
\frac{\partial V(x_p^t, y_p^t, c| \Theta,R)}{\partial U_{nm}^c} = \sum_{n}^{H} & \sum_{m}^{M} \max(0, 1 - | x_p^s - m|) \\
& \times \max(0, 1 - | y_p^s - n|)  
\end{split}
\label{eqn::stroi_partial_U}
\end{equation}

Moreover, during back propagation, we need to compute the gradient with respect to the projective transformation parameter vector $\Theta = [\theta_{1}, \theta_{2}, \theta_{3}; \theta_{4}, \theta_{5}, \theta_{6}; \theta_7, \theta_8, 1]$. Note that we set $\theta_9 = 1$ in (\ref{eqn::grid_generator}) hence we only need to calculate the gradient of $V$ with respect to eight projective parameters. In this way, RSN could also be updated to adjust the selected region. Although (\ref{eqn::stroi_bilinear}) may not be differentiable when $x_p^s = m$ or $y_p^s = n$, this seldom happens in practice because the possibility that the calculated $x_p^s$ or $y_p^s$ are integers is extremely low. We empirically set the gradients at these points to be $0$ as their effect on the back propagation process is negligible. 

We consider $\theta_7$ as examples due to space limitations and similar derivative can be computed for other parameters $\theta_i (i = 2,\dots,6, 8)$ respectively (See Appendix for a complete derivation). Denote $V(x_p^t, y_p^t, c| \Theta,R)$ as $V_p$ for simplicity, after applying the chain rule for the differentiable points:

\begin{equation}
\begin{split}
\frac{\partial V_p}{\partial \theta_{7}} = -\frac{x_p^t}{z_p^s} (\frac{\partial V_p}{\partial x_{p}^s}x_p^s  + \frac{\partial V_p}{\partial y_{p}^s}y_p^s) 
\end{split}
\label{eqn::stroi_partial_theta}
\end{equation}
where 
\begin{equation}
\small
\begin{split}
 \frac{\partial V_p}{\partial x_p^s} & = \sum_{n}^{H} \sum_{m}^{M} U_{nm}^c \max(0, 1 - | y_p^s - n|) \eta (x_p^s - m) \\
& \quad \eta (x_p^s - m) = 
\begin{cases}
0,  \quad \text{ if } |m - x_p^s| \ge 1 \\
1,  \quad \text{ if } m > x_p^s \\
-1, \quad \text{ if } m < x_p^s
\end{cases}
\end{split}
\label{eqn::stroi_partial_x}
\end{equation}
It is worth noting that $(x_p^t, y_p^t)$ are normalized coordinates in range $[-1, 1]$ so that they can to be scaled with respect to $w$ and $h$ with start position $(w_0, h_0)$. 

\subsubsection{Gating Network}

\begin{figure}[tbh]
\centering{
\includegraphics[width=0.45\columnwidth]{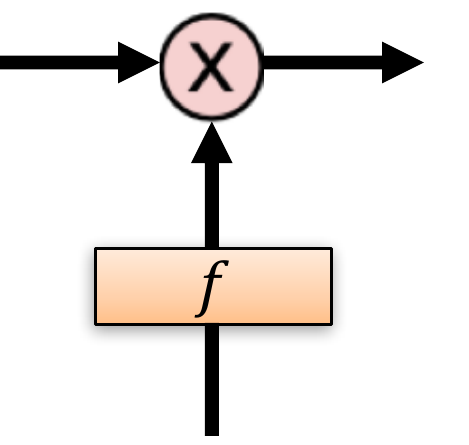}}
\caption{Design of the gating network. $f$ denotes the non-negative gate function (\textit{i.e.} sigmoid)}
\label{fig::gating_function}
\end{figure}

The gating network, which serves as a soft regionlet selector, is used to assign regionlets with different weights and generate regionlet feature representation. We design a simple gating network using a fully connected layer with \texttt{sigmoid} activation. The output values of the gating network are in the range $[0, 1]$.
Given the output feature maps $V(x_p^t, y_p^t, c| \Theta,R)$ described above, we use a fully connected layer to generate the same number of outputs as feature maps $V(x_p^t, y_p^t, c| \Theta,R)$, which is followed by an activation layer \texttt{sigmoid} to generate the corresponding weight respectively. The final feature representation is generated by the product of feature maps $V(x_p^t, y_p^t, c| \Theta,R)$ and their corresponding weights. 

\subsubsection{Regionlet Pool Construction}
Object deformations may occur at different scales. For instance, deformation could be caused by different body parts in person detection. Same number of regionlets (size $H\times W$) learned from small selected region have higher extraction density, which may lead to non-compact regionlet representation. In order to learn a \emph{compact} and \emph{efficient} regionlet representation, we further perform the pooling (\textit{i.e.} max/ave) operation over the feature maps $V(x_p^t, y_p^t, c| \Theta,R)$ of size ($H \times W$). 

We reap two benefits from the pool operation: (1) Regionlet representation is compact (small size). (2) Regionlets learned from different size of selected regions are able to represent such regions in an efficient way, and handle object deformations at different scales.

\section{Relations to Recent Works}\label{sec:related_discussion}

We review the traditional regionlet-based approach and its DNP extension in Section~\ref{sec:regionletsReview}. Besides this, our deep regionlet approach is related to some recently reported object detection works in different aspects. In this section, we discuss both similarities and differences in detail.

\subsection{Spatial Transform Networks} Jaderberg \textit{et al}.~\cite{Jaderberg_JSZK_NIPS2015} first proposed the spatial transformer module to provide spatial transformation capabilities into a deep neural network. It only learns \emph{one global parametric transformation} (scaling, rotations as well as affine transformation). Such learning is known to be difficult to apply on semi-dense vision tasks (\textit{e.g.}, object detection) and the transformation is on the entire feature map, which means the transformation is applied identically across all the regions in the feature map.

The RSN learns a set of projective transformation and each transformation can be considered as the localization network in~\cite{Jaderberg_JSZK_NIPS2015}.
However, our regionlet learning is different from image sampling~\cite{Jaderberg_JSZK_NIPS2015} method as it adopts a region-based spatial transformation and feature wrapping. By learning the transformation locally in the detection bounding box, the proposed method provides the flexibility of learning a compact, efficient feature representation of objects with variable shape and deformable parts. 

\subsection{Deformable Part Model and its deep learning extensions}
Deformable part models~\cite{Felzenszwalb_FGM_CVPR2010} explicitly represent spatial deformations of object parts via latent variables. A root filter is learned to model the global appearance of the objects, while the part filters are designed to describe the local parts in the objects. However, DPM is a shallow model and the training process involves heuristic choices to select components and part sizes, making end-to-end training inefficient.

Both works~\cite{Dai_DQXLZHW_ICCV2017,Mordan_MTCH_2017} extend the DPM with end-to-end training in deep CNNs. Motivated by DPM~\cite{Felzenszwalb_FGMR_TPAMI2010} to allow parts to slightly move around their reference position (partition of the initial regions), they share the similar idea of learning part offsets\footnote{\cite{Dai_DQXLZHW_ICCV2017} uses term offset while~\cite{Mordan_MTCH_2017} uses term displacement} to model the local element and pool the features at their corresponding locations after the shift. While~\cite{Dai_DQXLZHW_ICCV2017,Mordan_MTCH_2017} show promising improvements over other deep learning-based object detectors~\cite{Girshick_G_ICCV2015,Ren_RHGS_TPAMI2016}, it still lacks the flexibility of modeling non-rectangular objects with sharp shapes and deformable parts. 

It is noticeable that the deep regionlet learning module in the proposed method on the selected region is a generalization of Deformable RoI Pooling in~\cite{Dai_DQXLZHW_ICCV2017,Mordan_MTCH_2017}. First, we generalize the selected region to be non-rectangular by learning the projective transformation parameters. Such non-rectangular regions could provide the capabilities for \emph{scaling}, \emph{shifting} and \emph{rotation} around the original reference region. If we only enforce the RSN to learn the shift, our regionlet learning mechanism would degenerate to similar deformable RoI pooling as in~\cite{Dai_DQXLZHW_ICCV2017,Mordan_MTCH_2017}

\subsection{Spatial-based RoI Pooling}
Traditional spatial pyramid pooling~\cite{Lazebnik_LSP_CVPR2006} performs pooling over hand crafted regions at different scales. With the help of deep CNNs,~\cite{He_HZRS_ECCV20014} proposes to use spatial pyramid pooling in deep learning-based object detection. However, as the pooling regions over image pyramid still need to be carefully designed to learn the spatial layout of the pooling regions, the end-to-end training strategy is not well facilitated. 
In contrast, Our deep regionlet learning approach learns pooling regions end-to-end in deep CNNs. Moreover, the region selection step for learning regionlets accommodates different sizes of the regions. Hence, we are able to handle object deformations at different scales without generating the feature pyramid. 

\begin{table*}[!tbh]
\centering
\caption{Ablation study on the improvement of the proposed deep regionlets method over traditional \emph{regionlets}~\cite{Wang_WYZL_ICCV2013,Wang_WYZL_2015TPAMI} and its extension DNP~\cite{Zou_ZWSL_BMVC2014}. Results are reported on different network architecture backbones, \textit{i.e} AlexNet~\cite{Krizhevsky_KSH_2012NIPS}, VGG16~\cite{Simonyan_SZ_2014} ResNet-50~\cite{He_HZRS_CVPR16} and ResNet-101~\cite{He_HZRS_CVPR16}. Ours-A denotes RSN predicting affine transformation parameters.}
\label{tab::comparison_with_regionlet}
\centering{
\setlength{\tabcolsep}{12pt}
\resizebox{0.9\textwidth}{!}
{
\begin{tabular}{| c | c | c | c | }
\hline
Methods & Regionlet~\cite{Wang_WYZL_ICCV2013,Wang_WYZL_2015TPAMI}  &  DNP~\cite{Zou_ZWSL_BMVC2014}  &  Ours-A (AlexNet) \\
\hline
mAP@0.5($\%$) & 41.7 & 46.1 & 63.2  \\
\hline
Methods &  Ours-A (VGG16) & Ours-A (ResNet-50) & Ours-A (ResNet-101) \\
\hline
mAP@0.5($\%$) & 73.0 & 74.2 & 75.3 \\
\hline
\end{tabular}}
}
\end{table*}


\section{Experiments}
\label{sec:experiments}

\begin{table*}[tbh]
\centering
\caption{Ablation study of each component in deep regionlet approach. Output size $H\times W$ is set to $4\times 4$ for all the baselines. Ours-A denotes RSN predicting affine transformation parameters.}
\label{tab::each_component}
\centering{
\setlength{\tabcolsep}{10pt}
\resizebox{0.9\textwidth}{!}
{
\begin{tabular}{| c | c | c | c | c |  }
\hline
Methods & Global RSN  &  Offset-only RSN~\cite{Dai_DQXLZHW_ICCV2017,Mordan_MTCH_2017}  &  Non-gating & Ours-A \\
\hline
mAP@0.5($\%$) & 30.27 & 78.5 & 81.3 (+2.8) & 82.0 (+3.5)\\
\hline
\end{tabular}}
}
\end{table*}

\begin{table*}[tbh]
\centering
\caption{Results of ablation studies when a RSN selects different number of regions and regionlets are learned at different level of density.}
\label{tab::numOfRegions_regionletsDensity}
\centering{
\setlength{\tabcolsep}{10pt}
\resizebox{0.9\textwidth}{!}{
\begin{tabular}[b]{|c|*{5}{c|}}\hline
\backslashbox{$\#$ of Regions}{Regionlets Density}
& $2\times 2$ & $3\times 3$ & $4\times 4$& $5 \times 5$& $6\times 6$ \\ 
\hline
$4(2\times 2)$ regions &  78.0  & 79.2  &   79.9  & 80.2    & 80.3 \\
\hline
$9 (3\times 3)$ regions&  79.6  & 80.3  &   80.9  &  81.5 & 81.3 \\
\hline
$16 (4\times 4)$ regions& 80.0  & 81.0  & 82.0  & 81.6  &   80.8 \\
\hline
\end{tabular}}
}
\end{table*}

In this section, we present comprehensive experimental results of the proposed approach on two challenging benchmark datasets: PASCAL VOC~\cite{Everingham_EGWWZ_IJCV2010} and MS-COCO~\cite{Lin_LMBHPRDZ_ECCV2014}. 
There are in total $20$ categories of objects in PASCAL VOC~\cite{Everingham_EGWWZ_IJCV2010} dataset, which includes rigid objects such as cars and deformable objects like cats. We follow the common settings used in~\cite{Ren_RHGS_TPAMI2016,Bodla_BSCD_ICCV2017,Dai_DLHS_NIPS2016,Girshick_G_ICCV2015} to make extensive comparsions. 
More specifically, we train our deep model on (1) VOC $2007$ \texttt{trainval} and (2) union of VOC $2007$ \texttt{trainval} and VOC $2012$ \texttt{trainval} and evaluate on the VOC $2007$ \texttt{test} set. We also report results on VOC $2012$ \texttt{test} set with the model trained on the VOC $2007$ \texttt{trainvaltest} and VOC $2012$ \texttt{trainval}. In addition, we report results on the VOC $2007$ \texttt{test} split for ablation analysis. Mean average precision (mAP) is reported throughout all the experiments on PASCAL VOC. 

MS-COCO~\cite{Lin_LMBHPRDZ_ECCV2014} is a widely used challenging dataset, which contains $80$ object categories. Following the official settings in COCO website\footnote{\url{http://cocodataset.org/\#detections-challenge2017}}, we use the COCO 2017 \texttt{trainval} split (union of $135$k images from \texttt{train} split and $5$k images from \texttt{val} split) for training. We report the COCO-style average precision (mmAP) on \texttt{test-dev} $2017$ split, which requires evaluation from the MS-COCO server\footnote{The updated settings (2017) are different from the previous settings (2016, 2015) in~\cite{Bodla_BSCD_ICCV2017,Lin_LGGHD_ICCV2017,Dai_DQXLZHW_ICCV2017,Dai_DLHS_NIPS2016,Lin_LGGHD_ICCV2017}, as it includes different train/val sets.} for testing. 

For the base network, we use both VGG-16~\cite{Simonyan_SZ_2014} and ResNet-101~\cite{He_HZRS_CVPR16} to demonstrate the generalization of our approach regardless of which network backbone we use. The \emph{\'{a} trous} algorithm~\cite{Long_LSD_CVPR2015,Mallat_M_1999} is adopted in stage $5$ of ResNet-101. Following the suggested settings in~\cite{Dai_DLHS_NIPS2016,Dai_DQXLZHW_ICCV2017}, we also set the pooling size to $7$ by changing the conv5 stage's effective stride from $32$ to $16$ to increase the feature map resolution. In addition, the first convolution layer with stride $2$ in the conv5 stage is modified to $1$. Both backbone networks are intialized with the pre-trained ImageNet~\cite{He_HZRS_CVPR16,Krizhevsky_KSH_2012NIPS} model. 

In the following sections, we report the results of a series of ablation experiments to understand the behavior of the proposed deep regionlet approach. Furthermore, we present comparisons with state-of-the-art detectors~\cite{Ren_RHGS_TPAMI2016,Dai_DLHS_NIPS2016,Dai_DQXLZHW_ICCV2017,He_HGDG_ICCV2017,Lin_LGGHD_ICCV2017,Lin_LDGHHB_CVPR2017} on both PASCAL VOC~\cite{Everingham_EGWWZ_IJCV2010} and MS COCO~\cite{Lin_LMBHPRDZ_ECCV2014} datasets.

\subsection{Ablation Study on PASCAL VOC}\label{sec::ablation_study}

In this section, we comprehensively evaluate the proposed deep regionlets method on the PASCAL VOC~\cite{Everingham_EGWWZ_IJCV2010} detection benchmark to understand its behavior. Unless otherwise stated, all ablation studies are performed with RSN predicting the affine transformation parameters in the proposed approach. 

\subsubsection{Comparison with the conventional Regionlets detection schema}

We first evaluate how the deep regionlets framework improves over the conventional \emph{Regionlets} detection schema~\cite{Wang_WYZL_ICCV2013,Wang_WYZL_2015TPAMI} and its DNP extension~\cite{Zou_ZWSL_BMVC2014}. It is noted that DNP~\cite{Zou_ZWSL_BMVC2014} first attempted to utilize the deep features using AlexNet~\cite{Krizhevsky_KSH_2012NIPS}. In order to draw a fair comparison, we train our model on VOC $2007$ \texttt{trainval} and evaluate on the VOC $2007$ \texttt{test} set using AlexNet~\cite{Krizhevsky_KSH_2012NIPS}. The shorter side of image is set to be $600$ pixels and training is performed for $60$k iterations with single mini-batch size on $1$ GPU. The learning rate is set at $10^{-3}$ for the first $40$k iterations and $10^{-4}$ for the rest $20$k iterations. 

Results of improvements over traditional \emph{regionlets}~\cite{Wang_WYZL_ICCV2013,Wang_WYZL_2015TPAMI} and DNP~\cite{Zou_ZWSL_BMVC2014} are shown in Table~\ref{tab::comparison_with_regionlet}. First, although DNP~\cite{Zou_ZWSL_BMVC2014} improved over traditional \emph{regionlet}~\cite{Wang_WYZL_ICCV2013,Wang_WYZL_2015TPAMI} by almost $5\%$ with the help of deep features, our approach provide significant improvements over both traditional regionlet~\cite{Wang_WYZL_ICCV2013,Wang_WYZL_2015TPAMI} and DNP~\cite{Zou_ZWSL_BMVC2014} (more than $\mathbf{20\%}$ in terms of mAP) due to the power of end-to-end trainable framework. Second, the mAP can be significantly increased by using deeper and more powerful networks like ResNet-50 and ResNet-101~\cite{He_HZRS_CVPR16}. All these observations support the effectiveness of the integration of traditional regionlet method into end-to-end trainable deep learning framework.  

\subsubsection{Ablation study on each component} 

Next, we investigate the proposed approach to understand each component and its behavior. For a fair comparison, we adopt ResNet-101 as the backbone network for ablation studies. We train our model on the union set of VOC $2007+2012$ \texttt{trainval} as well as their horizontal
flip and evaluate on the VOC $2007$ \texttt{test} set. The shorter side of image is set at $600$ pixels, as suggested in~\cite{Girshick_G_ICCV2015,Ren_RHGS_TPAMI2016,Dai_DLHS_NIPS2016,Dai_DQXLZHW_ICCV2017}. The training is performed for $60$k iterations with an effective mini-batch size of $4$ on $4$ GPUs, where the learning rate is set at $10^{-3}$ for the first $40$k iterations and at $10^{-4}$ for the rest $20$k iterations. We investigate the proposed approach to understand each component (1) RSN, (2) Deep regionlet learning and (3) Soft regionlet selection by comparing it with several baselines: 
\begin{enumerate}
  \item Global RSN. RSN only selects one global region and it is initialized as identity affine transformation (\textit{i.e.} $\Theta_0^\ast = [1, 0, 0; 0, 1, 0]$). This is equivalent to global regionlet learning within the RoI.
  \item Offset-only RSN. We set the RSN to only learn the offset by enforcing $\theta_1, \theta_2, \theta_4, \theta_5$ (in affine parameters) to not change during the training process. In this way, the RSN only selects the rectangular region with offsets to the initialized region. This baseline is similar to the Deformable RoI Pooling in ~\cite{Dai_DQXLZHW_ICCV2017} and ~\cite{Mordan_MTCH_2017}.
  \item Non-gating selection: deep regionlet without soft selection. No soft regionlet selection is performed after the regionlet learning. In this case, each regionlet learned has the same contribution to the final feature representation.
\end{enumerate}

Results are shown in Table~\ref{tab::each_component}. First, when the RSN only selects one global region, the RSN reduces to the single localization network~\cite{Jaderberg_JSZK_NIPS2015}. In this case, regionlets are extracted in a global manner. It is interesting to note that selecting only one region by the RSN is able to converge, which is different from~\cite{Ren_RHGS_TPAMI2016,Dai_DLHS_NIPS2016}. However, the performance is extremely poor. This is because no discriminative regionlets could be explicitly learned within the region. More importantly, the results clearly demonstrate that the RSN is \textbf{\emph{indispensable}} to the deep regionlet approach. 

Moreover, offset-only RSN could be viewed as similar to deformable RoI pooling in~\cite{Dai_DQXLZHW_ICCV2017,Mordan_MTCH_2017}. These methods all learn the offset of the rectangle region with respect to its reference position, which lead to improvement over~\cite{Ren_RHGS_TPAMI2016}. However, non-gating selection outperforms offset-only RSN by $2.8\%$ with selecting non-rectangular region. The improvement demonstrates that non-rectangular region selection could provide more flexibility around the original reference region, thus could better model non-rectangular objects with sharp shapes and deformable parts. Last but not least, by using the gate function to perform soft regionlet selection, the performance can be further improved by $0.7\%$.

Next, we present ablation studies on the following questions in order to understand more deeply the RSN and regionlet learning modules. We report the results where the RSN predicts the affine transformation parameters: 
\begin{enumerate}
\item How many regions should we learn by RSN? 
\item How many regionlets should we learn in one selected region (density is of size $H\times W$)? 
\end{enumerate}

\subsubsection{How many regions should RSN learn?} 
We investigate how the detection performance varies when different number of regions are selected by the RSN. All the regions are initialized as described in Section~\ref{sec::region_selection} without any overlap among the regions. Without loss of generality, we report results for $4 ( 2\times 2)$, $9 (3 \times 3)$ and $16 (4 \times 4)$ regions in Table~\ref{tab::numOfRegions_regionletsDensity}. We observe that the mean AP increases when the number of selected regions is increased from $4(2 \times 2)$ to $9 (3 \times 3)$ for fixed regionlets learning number, but gets saturated with $16(4 \times 4)$ selected regions. 

\subsubsection{How many regionlets should we learn in one selected region?} 
Next, we investigate how the detection performance varies when different number of regionlets are learned in one selected region by varying $H$ and $W$. Without loss of generality, we set $H = W$ throughout our experiments and vary the $H$ from $2$ to $6$. In Table~\ref{tab::numOfRegions_regionletsDensity}, we report results when we set the number of regionlets at $4 (2 \times 2)$, $9 (3 \times 3)$, $16 (4 \times 4)$, $25 (5 \times 5)$, $36 (6 \times 6)$ before the regionlet pooling construction. 

First, it is observed that increasing the number of regionlets from $4(2\times 2)$ to $25 (5\times 5)$ results in improved performance. As more regionlets are learned from one region, more spatial and shape information from objects could be learned. The proposed approach could achieve the best performance when regionlets are extracted at $16 (4\times 4)$ or $25 (5 \times 5)$ density level. It is also interesting to note that when the density increases from $25(5 \times 5)$ to $36 (6 \times 6)$, the performance degrades slightly. When the regionlets are learned at a very high density level, some redundant spatial information may be learned without being useful for detection, thus affecting the region proposal-based decision to be made. Throughout all the experiments in the paper, we report the results from $16$ selected regions from RSN and set output size $H \times W = 4 \times 4$. 

\begin{table*}[tbh]
\caption{Detection results on PASCAL VOC$2007$ using VGG16 as backbone architecture. Training data: "07": VOC$2007$ \texttt{trainval}, "07 + 12": union set of VOC$2007$ and VOC$2012$ \texttt{trainval}. Ours-A(Ours-P)$^\S$ denotes applying the soft-NMS~\cite{Bodla_BSCD_ICCV2017} in the test stage.}
\label{tab::VOC2007_VGG16}
\centering{
\setlength{\tabcolsep}{11pt}
\resizebox{0.9\textwidth}{!}
{
\begin{tabular}{ c | c | c | c | c   }
\hline
Methods &training data & mAP@0.5($\%$) &  training data & mAP@0.5($\%$) \\
\hline
Regionlet~\cite{Wang_WYZL_ICCV2013} & 07 &  41.7 &  07 + 12 & N/A \\
\hline 
DNP~\cite{Zou_ZWSL_BMVC2014} & 07 & 46.1 & 07 + 12 & N/A \\
\hline
\hline
Faster R-CNN~\cite{Ren_RHGS_TPAMI2016} & 07 & 70.0 & 07 + 12 & 73.2 \\
\hline
R-FCN~\cite{Dai_DLHS_NIPS2016} & 07 & 69.6  & 07 + 12 & 76.6 \\
\hline
SSD512~\cite{Liu_LAESRFB_CVPR2016} & 07 & 71.6 &  07 + 12 & 76.8 \\
\hline
Soft-NMS~\cite{Bodla_BSCD_ICCV2017} & 07 & 71.1 & 07 + 12 & 76.8 \\
\hline
\hline
Ours-A & 07 & 73.0 & 07 + 12 & 79.2 \\
\hline
Ours-P & 07 & 73.3 & 07 + 12 & 79.5 \\
\hline 
Ours-A$^\S$ & 07 & 73.8 & 07 + 12 & 80.1 \\
\hline
Ours-P$^\S$ & 07 & 73.9 & 07 + 12 & 80.3 \\
\hline
\end{tabular}}
}
\end{table*}

\subsection{Experiments on PASCAL VOC}

In this section, we present experimental results on the PASCAL VOC dataset and compare our results thoroughly with several methods described below: 
\begin{itemize}
\item Traditional regionlet method~\cite{Wang_WYZL_ICCV2013} and DNP~\cite{Zou_ZWSL_BMVC2014}.
\item Popular deep learning-based object detectors: Faster R-CNN~\cite{Ren_RHGS_TPAMI2016}, SSD~\cite{Liu_LAESRFB_CVPR2016}, R-FCN~\cite{Dai_DLHS_NIPS2016}, soft-NMS~\cite{Bodla_BSCD_ICCV2017}, DP-FCN~\cite{Mordan_MTCH_2017} and DF-RCNN/D-RFCN~\cite{Dai_DQXLZHW_ICCV2017}.
\item State-of-the-art deep learning-based object detectors: MLKP~\cite{Wang__WWGLZ_CVPR2018}, RefineDet~\cite{Zhang_ZWBLL_CVPR2018}, PAD~\cite{Zhao_ZLW_CVPR2018}, DES~\cite{Zhang_ZQXSWY_CVPR2018} and STDN~\cite{Zhou_ZNGHX_CVPR2018}, RFB-Net~\cite{Liu_LHW_ECCV2018}, PFPNet-R~\cite{Kim_KKSKK_ECCV2018}, C-FRCNN~\cite{Chen_CHT_ECCV2018}, DFPN-R~\cite{Kong_KSHL_ECCV2018}.
\end{itemize}

\subsubsection{PASCAL VOC 2007}

\begin{table*}[htb]
\caption{Detection results on PASCAL VOC$2007$ test set. For fair comparison, we only list the results of single model without multi-scale training/testing, ensemble, iterative bounding box regression or additional segmentation label. Training data: union set of VOC $2007$ and $2012$ \texttt{trainval}. $^\ast$: the results are reported using new data augmentation trick. D-RFCN$^\dag$: this entry is obtained from~\cite{Dai_DQXLZHW_ICCV2017} using OHEM~\cite{Shrivastava_SGG_CVPR2016}. Ours-A(Ours-P)$^\S$ denotes we apply the soft-NMS~\cite{Bodla_BSCD_ICCV2017} in the test stage.}
\label{tab::VOC2007_ResNet101}
\centering{
\setlength{\tabcolsep}{11pt}
\resizebox{0.98\textwidth}{!}
{
\begin{tabular}{ c | c | c | c | c | c }
\hline
Methods & mAP@0.5($\%$) & Year & Methods & mAP@0.5($\%$) & Year \\
\hline
Fast R-CNN~\cite{Girshick_G_ICCV2015} & 70.0 & 2015 & Faster R-CNN~\cite{Ren_RHGS_TPAMI2016} &  78.1 & 2016 \\
\hline
OHEM~\cite{Shrivastava_SGG_CVPR2016} & 74.6 & 2016 & SSD$^\ast$~\cite{Liu_LAESRFB_CVPR2016} &  77.1  & 2016 \\
\hline 
\hline
ION~\cite{Bell_BZBG_CVPR2016} & 79.4 & 2016 & DP-FCN~\cite{Mordan_MTCH_2017} &  78.1 & 2017 \\
\hline
DF-RCNN(ROI Pooling)~\cite{Dai_DQXLZHW_ICCV2017} & 78.3 & 2017 & DF-RCNN~\cite{Dai_DQXLZHW_ICCV2017} & 79.3 & 2017 \\
\hline
D-RFCN(ROI Pooling)~\cite{Dai_DQXLZHW_ICCV2017} & 81.2 & 2017 & D-RFCN$^\dag$~\cite{Dai_DQXLZHW_ICCV2017} & 82.6 & 2017 \\
\hline 
\hline
MR-CNN~\cite{Gidaris_GK_ICCV2015} & 78.2 & 2015 & LocNet~\cite{Gidaris_GK_CVPR2016} & 78.4 & 2016 \\
\hline
DSSD~\cite{Fu_FLRTB_2017} & 78.6 & 2017 & PAD~\cite{Zhao_ZLW_CVPR2018} & 79.5 & 2018 \\
\hline
MLKP~\cite{Wang__WWGLZ_CVPR2018} & 80.6 & 2018 & DES$^\ast$~\cite{Zhang_ZQXSWY_CVPR2018} & 81.7 & 2018 \\
\hline
RefineDet~\cite{Zhang_ZWBLL_CVPR2018} & 80.0 & 2018 & STDN~\cite{Zhou_ZNGHX_CVPR2018} & 80.9 & 2018 \\
\hline 
RFB-Net~\cite{Liu_LHW_ECCV2018} & 82.2 & 2018 & PFPNet-R~\cite{Kim_KKSKK_ECCV2018} & 82.3 & 2018 \\
\hline 
C-FRCNN~\cite{Chen_CHT_ECCV2018} & 82.2 & 2018 & DFPN-R~\cite{Kong_KSHL_ECCV2018} & 82.4 & 2018 \\
\hline
\hline
Ours-A & 82.2 &  & Ours-A$^\S$ & 83.1 &  \\
\hline
Ours-P & \textbf{\emph{82.5}} &  & Ours-P$^\S$ & \textbf{83.3} &  \\
\hline
\end{tabular}}
}
\end{table*}

\begin{table*}[bht]
\caption{Complete Object Detection Results on PASCAL VOC $2007$ \texttt{test} set for each object category. Ours-A(Ours-P)$^\S$ denotes we apply the soft-NMS~\cite{Bodla_BSCD_ICCV2017} in the test stage.}
\label{tab::VOC2007_detail}
\centering{
\setlength{\tabcolsep}{1pt}
\resizebox{1\textwidth}{!}
{
\begin{tabular}{c|c|cccccccccccccccccccc}
Methods  & mAP & \rotatebox{90}{aero} & \rotatebox{90}{bike} & \rotatebox{90}{bird} & \rotatebox{90}{boat} & \rotatebox{90}{bottle} & \rotatebox{90}{bus} & \rotatebox{90}{car} & \rotatebox{90}{cat} & \rotatebox{90}{chair} & \rotatebox{90}{cow} & \rotatebox{90}{table} & \rotatebox{90}{dog} & \rotatebox{90}{horse} & \rotatebox{90}{mbike} & \rotatebox{90}{person} & \rotatebox{90}{plant} & \rotatebox{90}{sheep} & \rotatebox{90}{sofa} & \rotatebox{90}{train} & \rotatebox{90}{tv} \\
\hline     
Faster R-CNN~\cite{Ren_RHGS_TPAMI2016}
& 76.4 & 79.8 & 80.7 & 76.2 & 68.3 & 55.9 & 85.1 & 85.3 & 89.8 & 56.7 & 87.8 & 69.4 & 88.3 & 88.9 & 80.9 & 78.4 &  41.7 & 78.6 & 79.8 & 85.3 & 72.0\\
\hline
R-FCN~\cite{Dai_DLHS_NIPS2016} & 80.5 & 79.9 & 87.2 & 81.5 & 72.0 & 69.8 & 86.8 & 88.5 & 89.8 & 67.0 & 88.1 &  74.5 & 89.8 & \textbf{90.6} & 79.9 & 81.2 & 53.7 & 81.8 & 81.5 & 85.9 & 79.9\\
\hline
SSD~\cite{Liu_LAESRFB_CVPR2016} & 80.6 & 84.3 & 87.6 & 82.6 & 71.6 & 59.0 & 88.2 & 88.1 & 89.3 & 64.4 & 85.6 & 76.2 & 88.5 & 88.9 & 87.5 & 83.0 & 53.6 & 83.9 & 82.2 & 87.2 & 81.3 \\
\hline
DSSD~\cite{Fu_FLRTB_2017} & 81.5 & 86.6 & 86.2 & 82.6 & 74.9 & 62.5 & 89.0 & 88.7 & 88.8 & 65.2 & 87.0 & 78.7 & 88.2 & 89.0 & 87.5 & 83.7 & 51.1 & \emph{\textbf{86.3}} & 81.6 & 85.7 & \textbf{83.7} \\    
\hline 
RefineDet~\cite{Zhang_ZWBLL_CVPR2018} & 80.0 & 83.9 & 85.4 & 81.4 & 75.5 & 60.2 & 86.4 & 88.1 & 89.1 & 62.7 & 83.9 & 77.0 & 85.4 & 87.1 & 86.7 & 82.6 & 55.3 & 82.7 & 78.5 & \emph{\textbf{88.1}} & 79.4  \\
\hline
DCN~\cite{Dai_DQXLZHW_ICCV2017} & 81.4 & 83.9 & 85.4 & 80.1 & 75.9 & 68.8 & 88.4 & 88.6 & 89.2 & 68.0 & 87.2 & 75.5 & 89.5 & 89.0 & 86.3 & 84.8 & 54.1 & 85.2 & 82.6 & 86.2 & 80.3 \\
\hline
DFPN-R~\cite{Kong_KSHL_ECCV2018} & 82.4 & \textbf{92.0} & \emph{\textbf{88.2}} & 81.1 & 71.2 & 65.7 & 88.2 & 87.9 & \textbf{92.2} & 65.8 & 86.5 & \textbf{79.4} & \emph{\textbf{90.3}} & \emph{\textbf{90.4}} & \textbf{89.3} & \textbf{88.6} & \textbf{59.4} & \textbf{88.4} & 75.3 & \textbf{89.2} & 78.5 \\ 
\hline
C-FRCNN~\cite{Chen_CHT_ECCV2018} & 82.2 & 84.7 & \emph{\textbf{88.2}} & \emph{\textbf{83.1}} & 76.2 & 71.1 & 87.9 & 88.7 & 89.5 & 68.7 & \textbf{88.6} & 78.2 & 89.5 & 88.7 & 84.8 & 86.2 & 55.4 & 84.7 & 82.0 & 86.0 & 81.7 \\  
\hline
\hline
Ours-A & 82.2 & 83.9 & 87.6 & 79.7 & 76.3 & \emph{\textbf{71.4}} & 88.9 & 89.3 & 89.9 & 69.1 & 87.6 & 78.3 & 90.1 & 88.9 & 87.0 & 85.8 & 54.3 & 84.7 & 82.2 & 85.9 & 82.1\\
\hline
Ours-P & \textbf{\emph{82.5}} & 86.1 & 87.3 & 80.2 & 76.6 & 70.9 & \emph{\textbf{89.1}} & \emph{\textbf{89.5}} & 89.9 & 69.8 & 87.7 & \emph{\textbf{79.0}} & \textbf{90.4} & 89.0 & 86.4 & 86.0 & 54.9 & 84.8 & 83.0 & 87.4 & \emph{\textbf{82.2}} \\
\hline
Ours-A$^\S$ & 83.1 & \emph{\textbf{87.6}} & 88.1 & \emph{\textbf{83.1}} & \emph{\textbf{77.1}} & 71.2 & \emph{\textbf{89.1}} & \emph{\textbf{89.5}} & 89.8 & \textbf{70.9} & 87.8 & 78.4 & \emph{\textbf{90.3}} & 89.0 & 88.1 & 87.9 & 56.1 & 84.4 & \emph{\textbf{83.1}} & 87.8 & 81.7\\
\hline
Ours-P$^\S$ & \textbf{83.3} & 87.0 & \textbf{88.5} & \textbf{83.2} & \textbf{77.5} & \textbf{71.9} & \textbf{89.3} & \textbf{89.8} & \emph{\textbf{90.4}} & \emph{\textbf{70.4}} & \emph{\textbf{88.2}} & 78.7 & \textbf{90.4} & 89.4 & \emph{\textbf{88.7}} & \emph{\textbf{88.1}} & \emph{\textbf{57.0}} & 85.4 & \textbf{83.3} & 87.4 & 81.9 \\
\end{tabular}
}}
\end{table*}

\begin{table*}[bht]
\caption{Detection results on VOC$2012$ \texttt{test} set using training data "07++12": the union set of 2007 \texttt{trainvaltest} and 2012 \texttt{trainval}. SSD$^\ast$ denotes the new data augmentation. Ours-A(Ours-P)$^\S$ denotes we apply the soft-NMS~\cite{Bodla_BSCD_ICCV2017} in the test stage.}
\label{tab::VOC2012}
\centering{{}
\setlength{\tabcolsep}{8pt}
\resizebox{0.95\textwidth}{!}
{
\begin{tabular}{| c | c | c | c | c | c | c | }
\hline
Methods & FRCN~\cite{Ren_RHGS_TPAMI2016} &  YOLO9000~\cite{Redmon_RF_CVPR2017}  &  FRCN OHEM & DSSD~\cite{Fu_FLRTB_2017} &  SSD$^\ast$~\cite{Liu_LAESRFB_CVPR2016} & ION~\cite{Bell_BZBG_CVPR2016} \\
\hline
mAP@0.5($\%$) & 73.8 & 73.4 & 76.3 & 76.3 & 78.5 & 76.4 \\
\hline
Methods & R-FCN~\cite{Dai_DLHS_NIPS2016} & DP-FCN~\cite{Mordan_MTCH_2017} &  Ours-A & Ours-P & Ours-A$^\S$ &  Ours-P$^\S$ \\
\hline 
mAP@0.5($\%$)  & 77.6 & 79.5 & 80.4 & \emph{\textbf{80.6}} & 81.2 & \textbf{81.3} \\
\hline
\end{tabular}}
}
\end{table*}

We follow the standard settings as in~\cite{Ren_RHGS_TPAMI2016,Dai_DLHS_NIPS2016,Bodla_BSCD_ICCV2017,Dai_DQXLZHW_ICCV2017}  and report mAP scores using IoU thresholds at $0.5$. 

We first evaluate the proposed deep regionlet approach on the small training dataset VOC $2007$ \texttt{trainval}. For the training stage, we set the learning rate at $10^{-3}$ for the first $40$k iterations, then decrease it to $10^{-4}$ for the next $20$k iterations with single GPU. Next, we evaluate our approach on a large training dataset, created by merging VOC $2007$ and VOC $2012$ \texttt{trainval}. Due to using more training data, the number of iterations is increased. More specifically, we perform the same training process as described in Section~\ref{sec::ablation_study}. Moreover, we use $300$ RoIs at test stage from a single-scale image testing with setting the image shorter side to be $600$. It is noted that for a fair comparison, we do not deploy the multi-scale training/testing, ensemble, iterative bounding box regression, online hard example mining(OHEM)~\cite{Shrivastava_SGG_CVPR2016}, although it is shown in~\cite{Bodla_BSCD_ICCV2017,Dai_DQXLZHW_ICCV2017} that such enhancements could lead to additional performance boost. We report our results from RSN predicting both projective transformation parameters (Ours-P) and affine transformation parameters (Ours-A). 

\textbf{PASCAL VOC 2007 using VGG16 Backbone} The results on VOC$2007$ \texttt{test} using VGG$16$~\cite{Simonyan_SZ_2014} backbone are summarized in Table~\ref{tab::VOC2007_VGG16}. We first compare with traditional regionlet method~\cite{Wang_WYZL_ICCV2013}, DNP~\cite{Zou_ZWSL_BMVC2014} and several popular object detectors~\cite{Ren_RHGS_TPAMI2016,Liu_LAESRFB_CVPR2016,Bodla_BSCD_ICCV2017} when training using small size dataset (VOC$2007$ \texttt{trainval}). Next, we evaluate our method as we increase the training dataset (union set of VOC $2007$ and $2012$ \texttt{trainval}). Exploiting the power of deep CNNs, the deep regionlet approach has significantly improved the detection performance over the traditional regionlet method~\cite{Wang_WYZL_ICCV2013} and DNP~\cite{Zou_ZWSL_BMVC2014}. We also observe that more data always helps. Moreover, it is encouraging that soft-NMS~\cite{Bodla_BSCD_ICCV2017} is only applied in the test stage without modification in the training stage, which could directly improve over~\cite{Ren_RHGS_TPAMI2016} by $1.1\%$. In summary, our method is consistently better than all the compared methods and the performance could be further improved if we replace NMS with soft-NMS~\cite{Bodla_BSCD_ICCV2017}. 

\textbf{PASCAL VOC 2007 using ResNet-101 Backbone} Next, we change the network backbone from VGG16~\cite{Simonyan_SZ_2014} to ResNet-$101$~\cite{He_HZRS_CVPR16} and present corresponding results in Table~\ref{tab::VOC2007_ResNet101}. This is also the common settings for evaluating deep learning-based object detectors. Besides basic object detection framework, Faster R-CNN~\cite{Ren_RHGS_TPAMI2016}, SSD~\cite{Liu_LAESRFB_CVPR2016}, R-FCN~\cite{Dai_DLHS_NIPS2016}, soft-NMS~\cite{Bodla_BSCD_ICCV2017}, we also compare with our direct competitors DF-RCNN/D-RFCN~\cite{Dai_DQXLZHW_ICCV2017} and DP-FCN~\cite{Mordan_MTCH_2017} as discussed in Section~\ref{sec:related_discussion}. In addition, we also compare with most recent object detectors, MLKP~\cite{Wang__WWGLZ_CVPR2018}, RefineDet~\cite{Zhang_ZWBLL_CVPR2018}, PAD~\cite{Zhao_ZLW_CVPR2018}, DES~\cite{Zhang_ZQXSWY_CVPR2018} and STDN~\cite{Zhou_ZNGHX_CVPR2018}~\footnote{We report the results from original papers under same settings. Some papers reported best results using different networks and settings.}. 

First, compared to the performance in Table~\ref{tab::VOC2007_VGG16} using VGG16~\cite{Simonyan_SZ_2014} backbone architecture, the mAP can be significantly increased by using deeper networks like ResNet-101~\cite{He_HZRS_CVPR16}. Second, we outperform DP-FCN~\cite{Mordan_MTCH_2017} and DF-RCNN~\cite{Dai_DQXLZHW_ICCV2017} by $4.4\%$ and $3.2\%$ respectively. This provides the empirical support that our deep regionlet learning method could be treated as a \emph{generalization} of Deformable RoI Pooling in~\cite{Dai_DQXLZHW_ICCV2017,Mordan_MTCH_2017}, as discussed in Section~\ref{sec:related_discussion}. Moreover, the results demonstrate that selecting \emph{non-rectangular} regions from our method provide more capabilities including \emph{scaling}, \emph{shifting} and \emph{rotation} to learn the feature representations.

\begin{table*}[tbh]
\caption{Object detection results on MS COCO $2017$ \texttt{test-dev} using ResNet-101~\cite{He_HZRS_CVPR16} as backbone acchitecture. Training data: union set of $2017$ \texttt{train} and 2017 \texttt{val} set. SSD$^\ast$, DSSD$^\ast$ denote the new data augmentation}
\label{tab::COCO_ResNet101}
\centering{
\setlength{\tabcolsep}{4pt}
\resizebox{0.95\textwidth}{!}
{
\begin{tabular}{| c | c | c | c | c | c | c |}
\hline
Methods & Training Data & mmAP $0.5$:$0.95$ & mAP @0.5 & mAP small & mAP medium & mAP large \\
\hline
Faster R-CNN~\cite{Ren_RHGS_TPAMI2016} &  trainval & 24.4 &  45.7 & 7.9 & 26.6 & 37.2 \\
\hline
SSD$^\ast$\cite{Liu_LAESRFB_CVPR2016} & trainval & 31.2 &  50.4 & 10.2 & 34.5 & 49.8 \\ 
\hline  
DSSD$^\ast$~\cite{Fu_FLRTB_2017} & trainval & 33.2 & 53.5 & 13.0 & 35.4 & 51.1 \\
\hline 
R-FCN~\cite{Dai_DLHS_NIPS2016} & trainval & 30.8 & 52.6 & 11.8 & 33.9 & 44.8 \\
\hline 
DF-RCNN~\cite{Dai_DQXLZHW_ICCV2017} & trainval & 33.1 & 50.3 & 11.6 & 34.9 & 51.2 \\
\hline
D-RFCN~\cite{Dai_DQXLZHW_ICCV2017} & trainval & 34.5 & 55.0 & 14.0 & 37.7 & 50.3 \\
\hline
CoupleNet~\cite{Zhu_ZZWZWL_ICCV2017} & trainval & 34.4 & 54.8 & 13.4 & 38.1 & 50.8 \\
\hline
RefineDet512~\cite{Zhang_ZWBLL_CVPR2018} & trainval & 36.4 & 57.5 & 16.6 & 39.9 & 51.4 \\
\hline
RelationNet~\cite{Hu_HGZDW_CVPR2018} & trainval & 39.0 & 58.6 & - & - & - \\
\hline 
Cascade-RCNN~\cite{Cai_CV_CVPR2018} & trainval & 42.7 & 62.1 & 23.7 & 45.5 & 55.2 \\
\hline
Mask R-CNN~\cite{He_HGDG_ICCV2017} & trainval & 38.2 & 59.6 & 19.8 & 40.2 & 48.8  \\
\hline 
RetinaNet800~\cite{Lin_LGGHD_ICCV2017} & trainval &  39.1 & 59.1 & 21.8 & 42.7 & 50.2 \\ 
\hline 
Ours-A & trainval & 39.3 & 59.8 & 21.7 & 43.7 &  50.9 \\
\hline 
Ours-P & trainval & 39.9 & 61.7 & 22.9 & 44.1 & 51.7 \\  
\hline
\end{tabular}}
}
\end{table*}

Furthermore, we compare the proposed deep regionlet approach with the most recent published methods, PAD~\cite{Zhao_ZLW_CVPR2018}, MLKP~\cite{Wang__WWGLZ_CVPR2018}, DES$^\ast$~\cite{Zhang_ZQXSWY_CVPR2018}, RefineDet~\cite{Zhang_ZWBLL_CVPR2018} and STDN~\cite{Zhou_ZNGHX_CVPR2018}, RFB-Net~\cite{Liu_LHW_ECCV2018}, PFPNet-R~\cite{Kim_KKSKK_ECCV2018}, C-FRCNN~\cite{Chen_CHT_ECCV2018}, DFPN-R~\cite{Kong_KSHL_ECCV2018}. It can be seen that our method performs better than all the recent published methods including DES$^\ast$~\cite{Zhang_ZQXSWY_CVPR2018} ($0.8\%$), which used new data augmentation trick described in SSD$^\ast$~\cite{Liu_LAESRFB_CVPR2016}. Such trick is proven to boost the performance as shown in~\cite{Liu_LAESRFB_CVPR2016,Zhang_ZQXSWY_CVPR2018,Fu_FLRTB_2017}. It is also noted that D-RFCN~\cite{Dai_DQXLZHW_ICCV2017} reported $82.6\%$ using OHEM~\cite{Shrivastava_SGG_CVPR2016} while we do not deploy OHEM. 
We achieve comparable result compared to D-RFCN~\cite{Dai_DQXLZHW_ICCV2017} that uses OHEM. In summary, 
our method achieves state-of-the-art performance on object detection task when using ResNet-$101$ as backbone network. Note that all the other results~\cite{Bell_BZBG_CVPR2016,Gidaris_GK_CVPR2016,Gidaris_GK_ICCV2015,Zhao_ZLW_CVPR2018,Wang__WWGLZ_CVPR2018,Zhang_ZWBLL_CVPR2018,Zhou_ZNGHX_CVPR2018,Liu_LHW_ECCV2018,Kim_KKSKK_ECCV2018,Kong_KSHL_ECCV2018,Chen_CHT_ECCV2018} are reported without using extra training data (\textit{i.e.} COCO data), multi-scale training/testing~\cite{Singh_SND_2018}, OHEM, ensemble or other post processing techniques.

\textbf{Complete Object Detection Results} We present the complete object detection results of the proposed deep regionlet method on the PASCAL VOC $2007$ \texttt{test} set. Other results are reported from either the updated model~\cite{Liu_LAESRFB_CVPR2016,Fu_FLRTB_2017}, the complete detection results reported in the paper~\cite{Kong_KSHL_ECCV2018,Chen_CHT_ECCV2018} or the official code provided by the authors with suggested settings~\cite{Ren_RHGS_TPAMI2016,Dai_DLHS_NIPS2016,Dai_DQXLZHW_ICCV2017,Zhang_ZWBLL_CVPR2018}~\footnote{We only list the methods, which either reported complete detection results on VOC $2007$ or the code is public available}. Note that we produce slightly lower performance $81.4\%$ than $82.6\%$ reported in~\cite{Dai_DQXLZHW_ICCV2017}. The difference may come from sampling order of the images from the training set. Keeping this in mind, it is observed that our method achieves best average precision on majority of all the $20$ classes in VOC $2007$ \texttt{test} set.  

\subsubsection{PASCAL VOC 2012}

We also present our results evaluated on VOC $2012$ \texttt{test} set in Table~\ref{tab::VOC2012}. We follow the same experimental settings as in~\cite{Dai_DLHS_NIPS2016,Ren_RHGS_TPAMI2016,Fu_FLRTB_2017,Liu_LAESRFB_CVPR2016,Mordan_MTCH_2017} and train our model using VOC"07++12", which consists of VOC $2007$ \texttt{trainvaltest} and VOC $2012$ \texttt{trainval} set. It can be seen that our method achieves state-of-the-art performance. In particular, we outperform DP-FCN~\cite{Mordan_MTCH_2017}, which further proves the generalization of our method over Deformable ROI Pooling in~\cite{Mordan_MTCH_2017}.

\subsection{Experiments on MS COCO}

In this section, we evaluate the proposed deep regionlet approach on the MS COCO~\cite{Lin_LMBHPRDZ_ECCV2014} dataset and compare with other state-of-the-art object detectors: Faster R-CNN~\cite{Ren_RHGS_TPAMI2016}, SSD~\cite{Liu_LAESRFB_CVPR2016}, R-FCN~\cite{Dai_DLHS_NIPS2016}, Deformable F-RCNN/R-FCN~\cite{Dai_DQXLZHW_ICCV2017}, Mask R-CNN~\cite{He_HGDG_ICCV2017}, RetinaNet~\cite{Lin_LGGHD_ICCV2017}. 

We adopt ResNet-101~\cite{He_HZRS_CVPR16} as the backbone architecture to ensure a fair comparison. Following the settings in~\cite{He_HGDG_ICCV2017,Dai_DQXLZHW_ICCV2017,Lin_LGGHD_ICCV2017,Dai_DLHS_NIPS2016}, we set the shorter edge of the image to $800$ pixels. Training is performed for $280$k iterations with effective mini-batch size $8$ on 8 GPUs. We first train the model with $10^{-3}$ learning rate for the first $160$k iterations, followed by learning rate $10^{-4}$ and $10^{-5}$ for another $80$k iterations and the next $40$k iterations. Five scales and three aspect ratios are used for anchors. We report results using either the released models or the code from the original authors. It is noted that we only deploy single-scale image training (no scale jitter) without the iterative bounding box average throughout all the experiments, although these enhancements could further boost performance. 

Table~\ref{tab::COCO_ResNet101} shows the results on 2017 \texttt{test-dev} set\footnote{MS COCO server does not accept 2016 and 2015 test-dev evaluation. As a result, we are not able to report results on $2016,2015$ test-dev set.}, which contains $20,288$ images. Compared with the baseline methods Faster R-CNN~\cite{Ren_RHGS_TPAMI2016}, R-FCN~\cite{Dai_DLHS_NIPS2016} and SSD~\cite{Liu_LAESRFB_CVPR2016}, both Deformable F-RCNN/R-FCN~\cite{Dai_DQXLZHW_ICCV2017} and our method provide significant improvements over~\cite{Ren_RHGS_TPAMI2016,Dai_DLHS_NIPS2016,Liu_LAESRFB_CVPR2016} (+$3.7\%$ and +$8.5\%$). Moreover, it can be seen that our method outperform Deformable F-RCNN/R-FCN~\cite{Dai_DQXLZHW_ICCV2017} by decent margin($\sim$$4\%$). This observation further supports that the proposed deep regionlet learning module could be treated as a \emph{generalization} of~\cite{Dai_DQXLZHW_ICCV2017,Mordan_MTCH_2017}, as discussed in Section~\ref{sec:related_discussion}. It is also noted although most recent state-of-the-art object detectors based on Mask R-CNN\footnote{Note~\cite{He_HGDG_ICCV2017} reported best result using ResNeXt-$101$-FPN~\cite{Xie_XGDTH_CVPR2017}, we only compare the results in~\cite{He_HGDG_ICCV2017} using ResNet-101~\cite{He_HZRS_CVPR16} backbone for fair comparison.}~\cite{He_HGDG_ICCV2017} utilize multi-task training with segmentation labels, we still improve over~\cite{He_HGDG_ICCV2017} by $1.1\%$.  
In addition, the main contribution focal loss in~\cite{Lin_LGGHD_ICCV2017}, which overcomes the obstacle caused by the imbalance of positive/nagetive samples, is complimentary to our method. We believe it can be applied in our method to further boost the performance. In summary, compared with Mask R-CNN~\cite{He_HGDG_ICCV2017}, RetinaNet\footnote{\cite{Lin_LGGHD_ICCV2017} reported best result using multi-scale training for 1.5$\times$ longer iterations, we only compare the results without scale jitter during training.}~\cite{Lin_LGGHD_ICCV2017} and other recent detectors~\cite{Zhang_ZWBLL_CVPR2018,Hu_HGZDW_CVPR2018,Cai_CV_CVPR2018,Zhu_ZZWZWL_ICCV2017}, our method achieves competitive performance on MS COCO when using ResNet-$101$ as the backbone network. 
\subsection{Complexity Analysis: Parameters and Speed}

In this section, we present the analysis on the speed and parameter of the proposed deep regionlets approach.

\textbf{Runtime Speed: }We evaluate the runtime of our approach and compare with other two-stage object detectors, Faster R-CNN~\cite{Ren_RHGS_TPAMI2016}, R-FCN~\cite{Dai_DLHS_NIPS2016} using the original Caffe~\cite{Jia_JEJSJRST_2014} implementation and ResNet-101 backbone with Batch Normalization(BN) layers for a fair comparison. The time is reported on single Nvidia TITAN X GPU including image resizing, network forward and post-processing. On average, Faster R-CNN~\cite{Ren_RHGS_TPAMI2016} takes $0.37$s and R-FCN~\cite{Dai_DLHS_NIPS2016} takes $0.24$s per image, while our method take $0.49$s per image. In addition, we also report the runtime for DF-RCNN/D-RFCN~\cite{Dai_DQXLZHW_ICCV2017} on the same machine configuration for reference purpose. DF-RCNN takes about $0.34$s and D-RFCN takes $0.25$s per image, note that DF-RCNN/D-RFCN~\cite{Dai_DQXLZHW_ICCV2017} uses different MXNet framework instead of Caffe and some python layers in RPN have been optimized with CUDA implementation. 

\textbf{Number of Parameters: } The RSN has three fully connected layers (First two connected layers have output size of $256$, last fully connected layer has the output size of $9$), giving about $5.28$M ( $16\times(1024\times256+256\times256+256\times9) $ ) parameters, while deep regionlet learning module and gating network do not introduce new parameters. According to~\cite{Ren_RHGS_TPAMI2016,Dai_DLHS_NIPS2016,Hu_HGZDW_CVPR2018,Zhao_ZLW_CVPR2018}, in total, Faster R-CNN has about $58.3$M parameters, R-FCN has about $56.4$M parameters. Therefore, the total number of parameters is about $63.6$M on top of Faster R-CNN framework. The increase in the number of parameters could be considered minor.


\section{Conclusion}
\label{sec:conclusion}

In this paper, we present a novel deep regionlet-based approach for object detection. The proposed region selection network can select \emph{non-rectangular} region within the detection bounding box, such that an object with rigid shape and deformable parts can be better modeled. We also design the deep regionlet learning module so that both the selected regions and the regionlets can be learned simultaneously. Moreover, the proposed system can be trained in a fully end-to-end manner without additional efforts. Finally, we extensively evaluate our approach on two detection benchmarks for generic object detection. Experimental results show competitive performance over state-of-the-art.

\ifCLASSOPTIONcompsoc
  \section*{Acknowledgments}
\else
  \section*{Acknowledgment}
\fi

This research is based upon work supported by the Intelligence Advanced Research Projects Activity (IARPA) via Department of Interior/Interior Business Center (DOI/IBC) contract number D17PC00345. The U.S. Government is authorized to reproduce and distribute reprints for Governmental purposes not withstanding any copyright annotation theron. Disclaimer: The views and conclusions contained herein are those of the authors and should not be interpreted as necessarily representing the official policies or endorsements, either expressed or implied of IARPA, DOI/IBC or the U.S. Government.​

\ifCLASSOPTIONcaptionsoff
  \newpage
\fi

\appendix\label{derivates}

We present the derivatives of the loss function with respect to the projective transformation parameters $\Theta$. Denote $V(x_p^t, y_p^t, c| \Theta,R)$ as $V_p$:  

\begin{equation}
\begin{split}
\frac{\partial V_p}{\partial \theta_{1}} & = \frac{\partial V_p}{\partial x_{p}^s} \frac{\partial x_p^s}{\partial \theta_{1}} = \frac{x_p^t}{z_p^s} \frac{\partial V_p}{\partial x_{p}^s} \\
\frac{\partial V_p}{\partial \theta_{2}} & = \frac{\partial V_p}{\partial x_{p}^s} \frac{\partial x_p^s}{\partial \theta_{2}} = \frac{y_p^t}{z_p^s} \frac{\partial V_p}{\partial x_{p}^s} \\
\frac{\partial V_p}{\partial \theta_{3}} & = \frac{\partial V_p}{\partial x_{p}^s} \frac{\partial x_p^s}{\partial \theta_{3}} = \frac{\partial V_p}{\partial x_{p}^s} \\
\frac{\partial V_p}{\partial \theta_{4}} & = \frac{\partial V_p}{\partial y_{p}^s} \frac{\partial y_p^s}{\partial \theta_{4}} = \frac{x_p^t}{z_p^s} \frac{\partial V_p}{\partial y_{p}^s} \\
\frac{\partial V_p}{\partial \theta_{5}} & = \frac{\partial V_p}{\partial y_{p}^s} \frac{\partial y_p^s}{\partial \theta_{5}} = \frac{y_p^t}{z_p^s} \frac{\partial V_p}{\partial y_{p}^s} \\
\frac{\partial V_p}{\partial \theta_{6}} & = \frac{\partial V_p}{\partial y_{p}^s} \frac{\partial y_p^s}{\partial \theta_{6}} = \frac{\partial V_p}{\partial y_{p}^s} \\
\frac{\partial V_p}{\partial \theta_{7}} & = \frac{\partial V_p}{\partial z_{p}^s} \frac{\partial z_p^s}{\partial \theta_{7}}  = (\frac{\partial V_p}{\partial x_{p}^s} \frac{\partial x_p^s}{\partial z_p^s} + \frac{\partial V_p}{\partial y_{p}^s} \frac{\partial y_p^s}{\partial z_p^s}) x_p^t  \\
 = - x_p^t & ( \frac{\theta_1 x_p^s + \theta_2 y_p^s + \theta_3}{{z_p^s}^2} \frac{\partial V_p}{\partial x_{p}^s} + \frac{\theta_4 x_p^s + \theta_5 y_p^s + \theta_6}{{z_p^s}^2} \frac{\partial V_p}{\partial y_{p}^s}) \\
 = -\frac{x_p^t}{z_p^s} & (\frac{\partial V_p}{\partial x_{p}^s}x_p^s  + \frac{\partial V_p}{\partial y_{p}^s}y_p^s) \\ 
 \frac{\partial V_p}{\partial \theta_{8}} & = \frac{\partial V_p}{\partial z_{p}^s} \frac{\partial z_p^s}{\partial \theta_{8}}  = (\frac{\partial V_p}{\partial x_{p}^s} \frac{\partial x_p^s}{\partial z_p^s} + \frac{\partial V_p}{\partial y_{p}^s} \frac{\partial y_p^s}{\partial z_p^s}) y_p^t  \\
 = - y_p^t & ( \frac{\theta_1 x_p^s + \theta_2 y_p^s + \theta_3}{{z_p^s}^2} \frac{\partial V_p}{\partial x_{p}^s} + \frac{\theta_4 x_p^s + \theta_5 y_p^s + \theta_6}{{z_p^s}^2} \frac{\partial V_p}{\partial y_{p}^s}) \\
 = -\frac{y_p^t}{z_p^s} & (\frac{\partial V_p}{\partial x_{p}^s}x_p^s  + \frac{\partial V_p}{\partial y_{p}^s}y_p^s) \\ 
\end{split}
\label{eqn::stroi_partial_theta}
\end{equation}
where 
\begin{equation}
\small
\begin{split}
 \frac{\partial V_p}{\partial x_p^s} & = \sum_{n}^{H} \sum_{m}^{M} U_{nm}^c \max(0, 1 - | y_p^s - n|) \eta (x_p^s - m) \\
 \frac{\partial V_p}{\partial y_p^s} & = \sum_{n}^{H} \sum_{m}^{M} U_{nm}^c \max(0, 1 - | x_p^s - m|) \eta (y_p^s - n) \\
& \quad \eta (x_p^s - m) = 
\begin{cases}
0,  \quad \text{ if } |m - x_p^s| \ge 1 \\
1,  \quad \text{ if } m > x_p^s \\
-1, \quad \text{ if } m < x_p^s
\end{cases} \\
& \quad \eta (y_p^s - n) = 
\begin{cases}
0,  \quad \text{ if } |n - y_p^s| \ge 1 \\
1,  \quad \text{ if } n > y_p^s \\
-1, \quad \text{ if } n < y_p^s
\end{cases} 
\end{split}
\label{eqn::stroi_partial_x}
\end{equation}



%
{\small
\bibliographystyle{ieee}
\bibliography{tpamibib}
}

%

\begin{IEEEbiography}[{\includegraphics[width=1in,height=1.25in,clip,keepaspectratio]{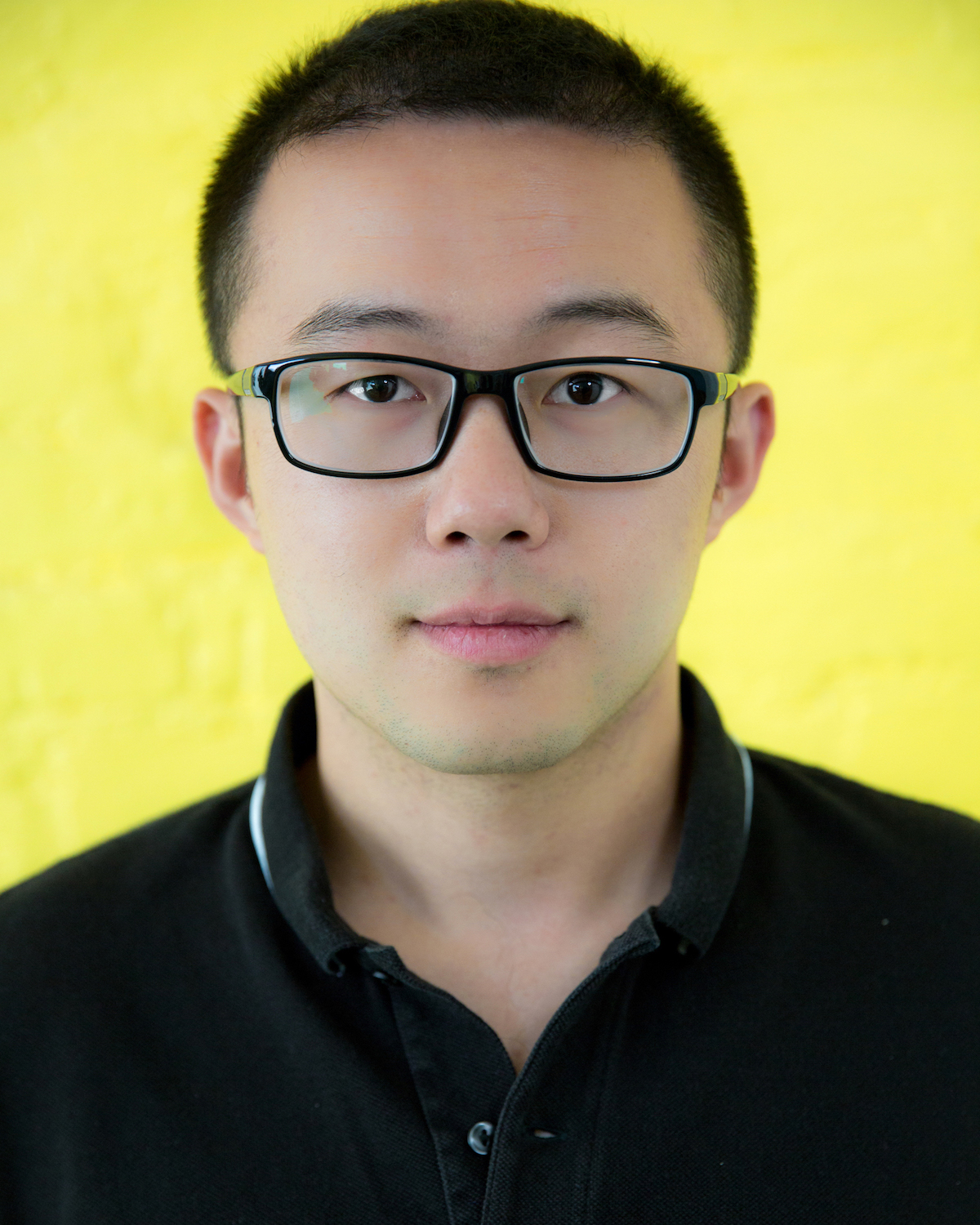}}]{Hongyu Xu}
received M.S. and Ph.D. degree in Electrical and Computer Engineering from University of Maryland, College Park in 2017 and 2019, respectively. He received the B.E. degree from the University of Science and Technology of China in 2012. He is currently a senior machine learning/computer vision engineer at Apple Inc. He is a former research intern with Snap Research (summer, fall 2017) and Palo Alto Research Center (PARC) (summer 2014). His research interests include object detection, deep learning, dictionary learning, face recognition, object classification, and domain adaptation.
\end{IEEEbiography}


\begin{IEEEbiography}[{\includegraphics[width=1in,height=1.25in,clip,keepaspectratio]{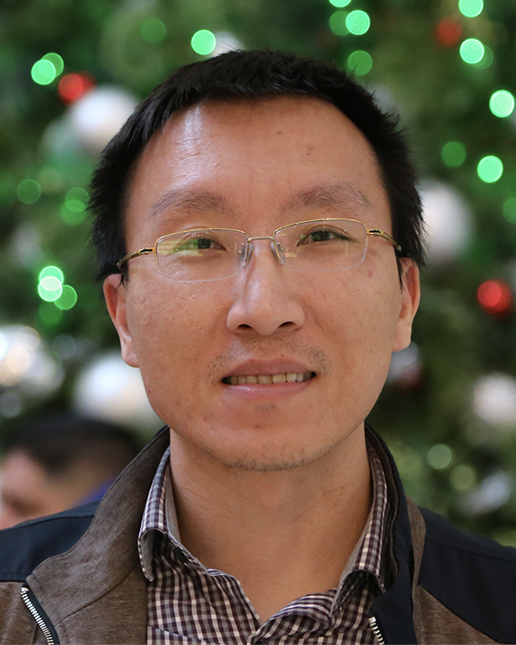}}]{Xutao Lv}
received his PhD degree from University of Missouri, Columbia in 2012. He is currently a Director in Intellifusion. Before joining Intellifusion, he worked as Senior Research Scientist in Snap Research since 2015 and worked as a Computer Scientist in SRI International from 2012. His research interests include computer vision and machine learning. 
\end{IEEEbiography}

\begin{IEEEbiography}[{\includegraphics[width=1in,height=1.25in,clip,keepaspectratio]{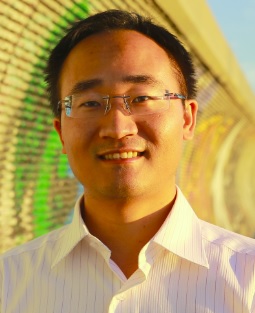}}]{Xiaoyu Wang} received his PhD degree in Electrical and Computer Engineering and MS in Statistics from University of Missouri, Columbia in 2012. He obtained his BE from University of Science and Technology of China in 2006. As a National Distinguished Expert, he is currently Co-founder and Chief Scientist of Intellifusion. He also serves as Guest Professor in the Chinese University of Hong Kong (Shenzhen) and
Shenzhen University. Before joining Intellifusion, he was a founding
member of Snap Research and served as Chair of Computer Vision
from 2015 to 2017. He was a Research Scientist at NEC Labs America
from 2012-2015. He served as area chair for WACV 2017. His team
was the runner-up winner of ImageNet object recognition challenge in
2013. His research interest is focused on Computer Vision and Machine
Learning.
\end{IEEEbiography}

\begin{IEEEbiography}[{\includegraphics[width=1in,height=1.25in,clip,keepaspectratio]{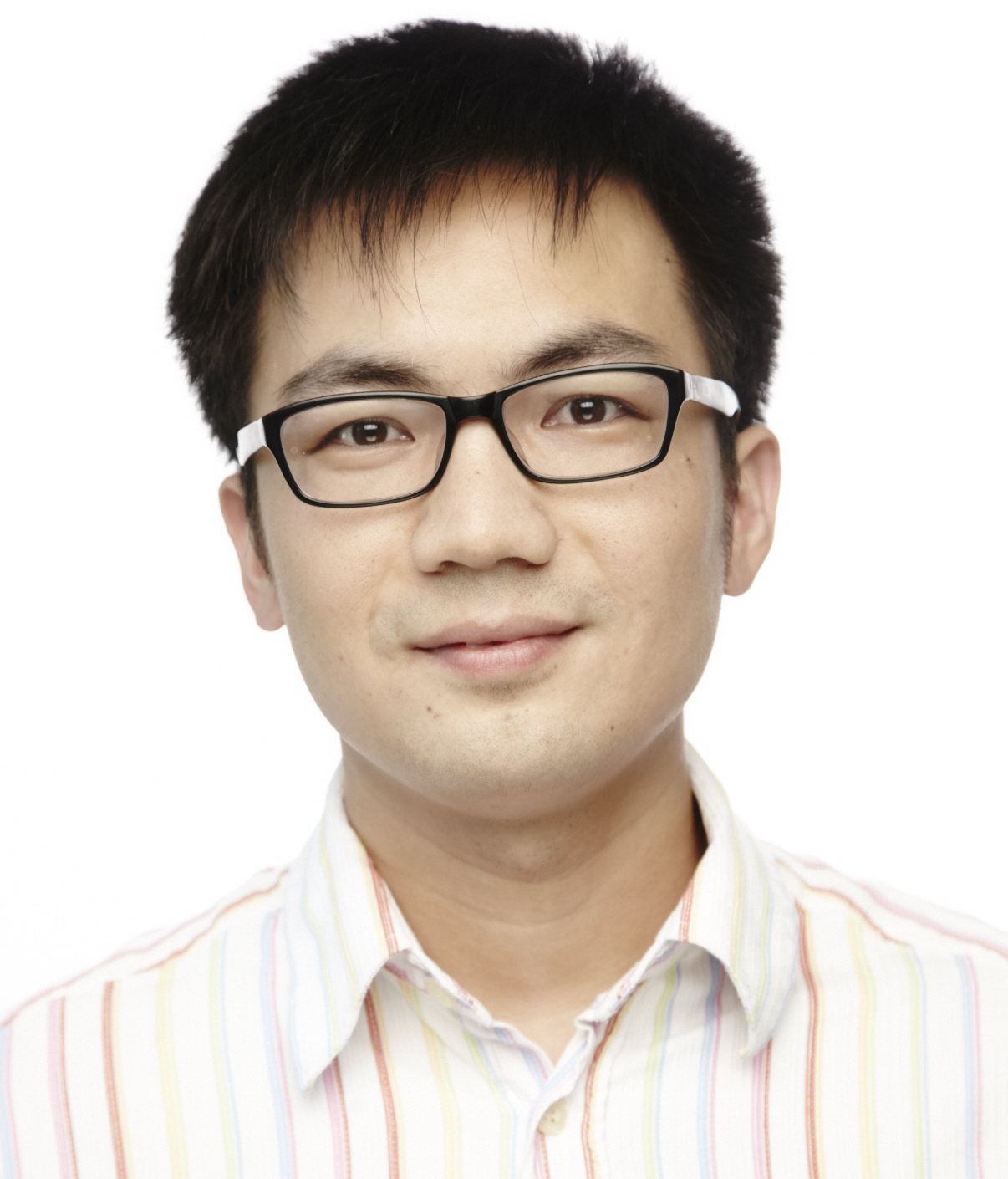}}]{Zhou Ren} is currently a Research Lead/Senior Researcher at Wormpex AI Research, USA. Before that, he was a Senior Research Scientist at Snap Inc. from 2016 to 2018. He received his Ph.D. from the Computer Science Department at University of California, Los Angeles and M.Eng. from Nanyang Technological University, Singapore, in 2016 and 2012 respectively. He is currently an Associate Editor of The Visual Computer Journal. His current research interests include computer vision, video analysis, multimodal understanding, etc. Zhou Ren was the recipient of the 2016 Best Paper Award from IEEE Transactions on Multimedia and the runner-up winner of the NIPS 2017 Adversarial Attack and Defense Competition. He was nominated to the CVPR 2017 Best Student Paper Award.
\end{IEEEbiography}

\begin{IEEEbiography}[{\includegraphics[width=1in,height=1.25in,clip,keepaspectratio]{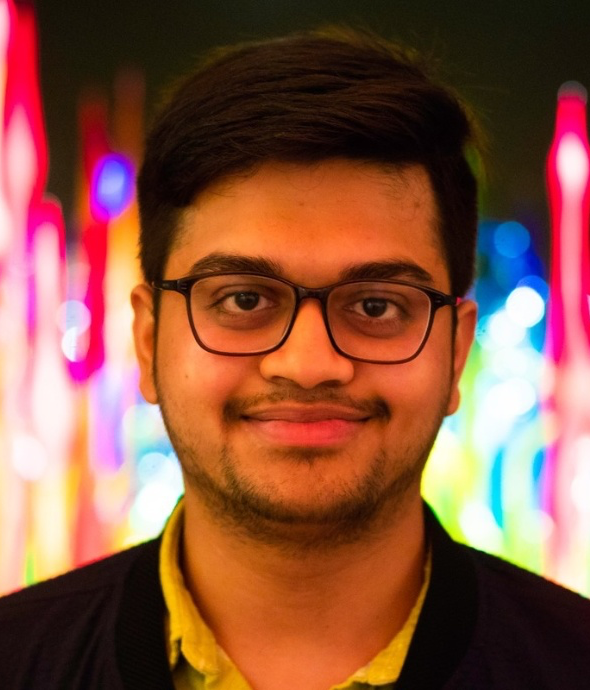}}]{Navaneeth Bodla} is a PhD candidate at the University of Maryland, College Park. He received the M.S degree in Computer Engineering from the University of Florida, Gainesville in 2014. He received the B.E degree from Birla Institute of Technology and Science, Pilani Hyderabad in 2012. He is a former research intern at Microsoft Research, Redmond (summer 2017, 2018) and Apple AI Research(Spring 2019). His research interests include object detection, deep learning, generative modeling and face recognition.
\end{IEEEbiography}

\begin{IEEEbiography}[{\includegraphics[width=1in,height=1.25in,clip,keepaspectratio]{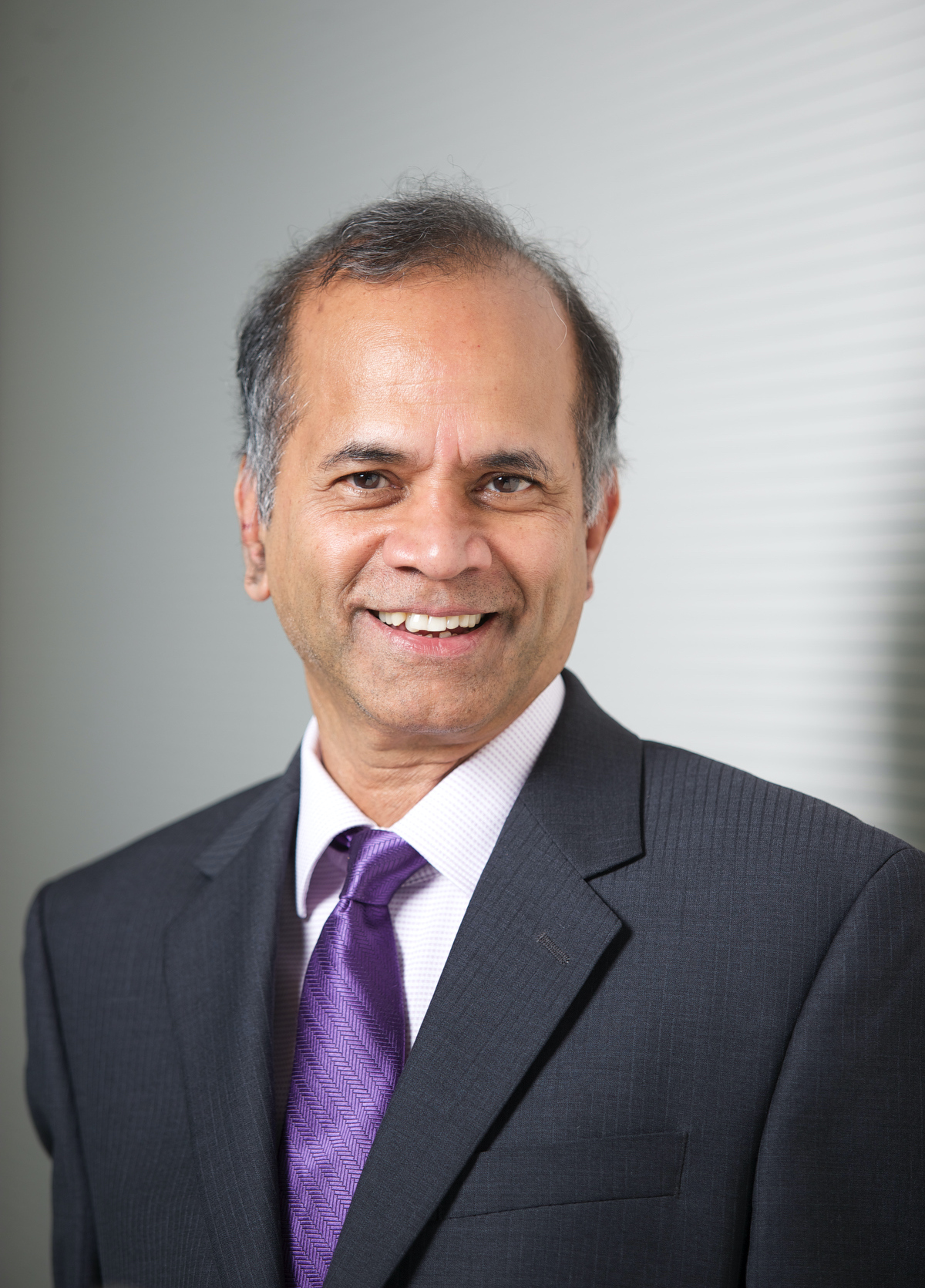}}]
{Prof. Rama Chellappa} is a Distinguished University Professor and a Minta Martin
Professor of Engineering in the Department of Electrical and Computer engineering
at the University of Maryland (UMD). At UMD, he is also affiliated with the Computer
Science Department, the University of Maryland Institute Advanced Computer Studies,
the Applied Mathematics and Scientific Computing program and the Center for Automation
Research. His current researcher interests are computer vision, pattern recognition and
machine intelligence. He has received numerous research, teaching, service, and
mentoring awards from the University of Southern California, University of Maryland,
IBM, IEEE Computer and Signal Processing Societies, the IEEE Biometrics Council and
the International Association of Pattern Recognition. He has been recognized as an
Outstanding Electrical Engineer by the ECE department at Purdue University and as a
Distinguished Alumni of the Indian Institute of Science. He is a Fellow of IEEE, IAPR,
OSA, AAAS, ACM, and AAAI and holds six patents.
\end{IEEEbiography}






\end{document}